\documentclass{article}

\usepackage{iclr2027_conference}
\usepackage{times}

\usepackage[T1]{fontenc}

\usepackage{amsfonts}
\usepackage{amssymb}
\usepackage{bm}
\usepackage{mathtools}

\newcommand{\vx}{\bm{x}}
\newcommand{\vy}{\bm{y}}
\newcommand{\ve}{\bm{e}}
\newcommand{\vu}{\bm{u}}
\newcommand{\vz}{\bm{z}}
\newcommand{\vg}{\bm{g}}
\newcommand{\veps}{\bm{\epsilon}}
\newcommand{\vxi}{\bm{\xi}}
\newcommand{\mI}{\bm{I}}
\usepackage{array}
\usepackage{booktabs}
\usepackage{multirow}
\usepackage{tabularx}
\usepackage{graphicx}
\usepackage{placeins}
\usepackage{hyperref}
\usepackage{url}
\usepackage{amsthm}
\usepackage{cleveref}
\usepackage{algorithmic}
\usepackage{enumitem}

\setlist[enumerate]{leftmargin=.5in}
\setlist[itemize]{leftmargin=.5in}

\newcounter{algorithm}
\newcommand{\algorithmcaption}[1]{%
  \refstepcounter{algorithm}%
  \par\noindent\textbf{Algorithm~\thealgorithm.}~#1\par
}
\crefname{algorithm}{Algorithm}{Algorithms}
\Crefname{algorithm}{Algorithm}{Algorithms}

\newtheorem{corollary}{Corollary}
\newtheorem{proposition}{Proposition}

\crefname{assumption}{Assumption}{Assumptions}
\Crefname{assumption}{Assumption}{Assumptions}

\theoremstyle{remark}

\crefname{hypothesis}{Hypothesis}{Hypotheses}
\Crefname{hypothesis}{Hypothesis}{Hypotheses}

\crefname{fact}{Fact}{Facts}
\Crefname{fact}{Fact}{Facts}

\usepackage{amsopn}

\title{FANO: Anchored Forward-Noise Optimization for Diffusion Inverse Problems}

\author{Shilong Han \quad Hongxia Wang\thanks{Corresponding author.} \\
College of Science, National University of Defense Technology \\
\texttt{hanshilong20@nudt.edu.cn, wanghongxia@nudt.edu.cn}}

\hypersetup{
  hidelinks,
  pdftitle={FANO: Anchored Forward-Noise Optimization for Diffusion Inverse Problems},
  pdfauthor={Shilong Han, Hongxia Wang}
}

\iclrfinalcopy

\begin{document}

\maketitle
\lhead{Preprint. Under review.}

\begin{figure}[t]
\centering
\includegraphics[width=\textwidth]{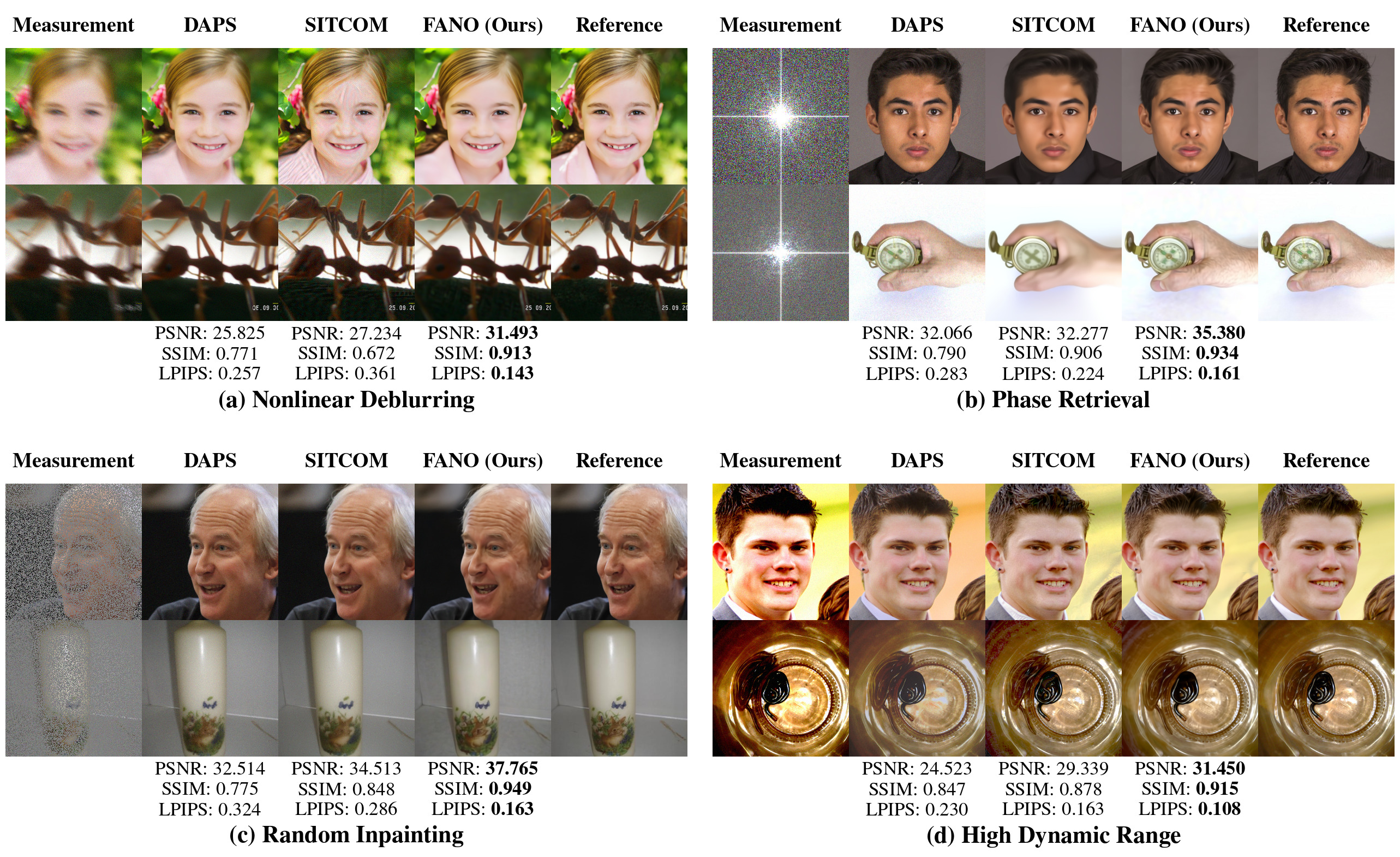}
\vspace{-0.5em}
\caption{Visual comparison on FFHQ and ImageNet across four inverse problems ($\sigma_y{=}0.05$): (a) nonlinear deblurring, (b) phase retrieval, (c) random inpainting, (d) high dynamic range (HDR). Each panel shows the measurement, DAPS, SITCOM, FANO, and the reference. Metrics below each panel are averaged over the two examples, with the best in \textbf{bold}; FANO recovers sharper structures and finer textures across all tasks, with the advantage most pronounced on nonlinear problems.}
\label{fig:teaser}
\end{figure}

\begin{abstract}
Pre-trained diffusion models have become a widely adopted zero-shot prior for image inverse problems, among which annealed decoupling methods achieve leading reconstruction performance on nonlinear tasks. We find, however, that annealed methods that directly optimize the denoiser input $\vx_t$ exhibit \emph{radial drift} in later layers: within-layer optimization drives the noise-component norm of $\vx_t$ away from its typical value, pushing the denoiser out of its high-performance regime and degrading reconstruction quality. To address this, we propose FANO (Anchored Forward-Noise Optimization). At each annealing layer, FANO fixes the previous denoiser estimate as an anchor and optimizes only the forward noise $\ve$, whose distribution is known, instead of $\vx_t$, thereby preserving the forward-diffusion structure of the denoiser input. The standard Gaussian prior on $\ve$ concentrates its norm near the typical value, and deviations degrade denoiser performance. We accordingly decompose $\ve$ in polar coordinates into a direction and a radial magnitude, and apply a soft potential to regulate the radius, keeping $\vx_t$ in the high-density region of the forward distribution. On multiple linear and nonlinear inverse problems, FANO matches the strongest current baselines on linear tasks and achieves consistent improvements on nonlinear tasks.
\end{abstract}

\section{Introduction}
\label{sec:introduction}

From a degraded measurement $\vy\in\mathbb R^m$, image inverse problems aim to recover an unknown image $\vx^\star\in\mathbb R^d$, with the two related through the forward measurement model $\vy=A(\vx^\star)+\vxi$, where $A:\mathbb R^d\to\mathbb R^m$ is a linear or nonlinear measurement operator and $\vxi\in\mathbb R^m$ denotes measurement noise. Owing to information loss ($m<d$ or noninvertible $A$) and noise, inverse problems are typically severely ill-posed and require prior information to constrain the solution space.

Pretrained diffusion models have become the most widely adopted generative priors for zero-shot image inverse problems \citep{ho2020denoising,song2021scorebased}. These methods directly use the denoiser $D_t$ provided by an unconditional diffusion model as an implicit image prior and incorporate measurement consistency into reverse sampling through the known $A$ at inference time, without retraining the model for each degradation type.

Among diffusion-prior methods for inverse problems (Section~\ref{sec:background}), annealed decoupling methods \citep{zhang2024daps,alkhouri2025sitcom,fastdips2026} decompose inference into a sequence of within-layer subproblems along an annealing schedule, achieving the strongest current reconstruction performance on nonlinear inverse problems. However, when analyzing SITCOM \citep{alkhouri2025sitcom}, an annealed method that directly optimizes the denoiser input, we find that the measurement residual continues to decrease in later annealing layers while reconstruction quality abruptly collapses (Figure~\ref{fig:diagnosis}(a,c)). Tracking the norm of the noise component in the denoiser input reveals a sharp increase in later layers (Figure~\ref{fig:diagnosis}(a)); controlled experiments show that this deviation is accompanied by degraded denoiser performance (Figure~\ref{fig:diagnosis}(b)). Moreover, the norm deviates more severely in the absence of measurement noise, ruling out the overfitting hypothesis. We refer to this phenomenon as \emph{radial drift}. Section~\ref{subsec:diagnosis} presents the full analysis.

\begin{figure*}[t]
\centering
\includegraphics[width=\textwidth]{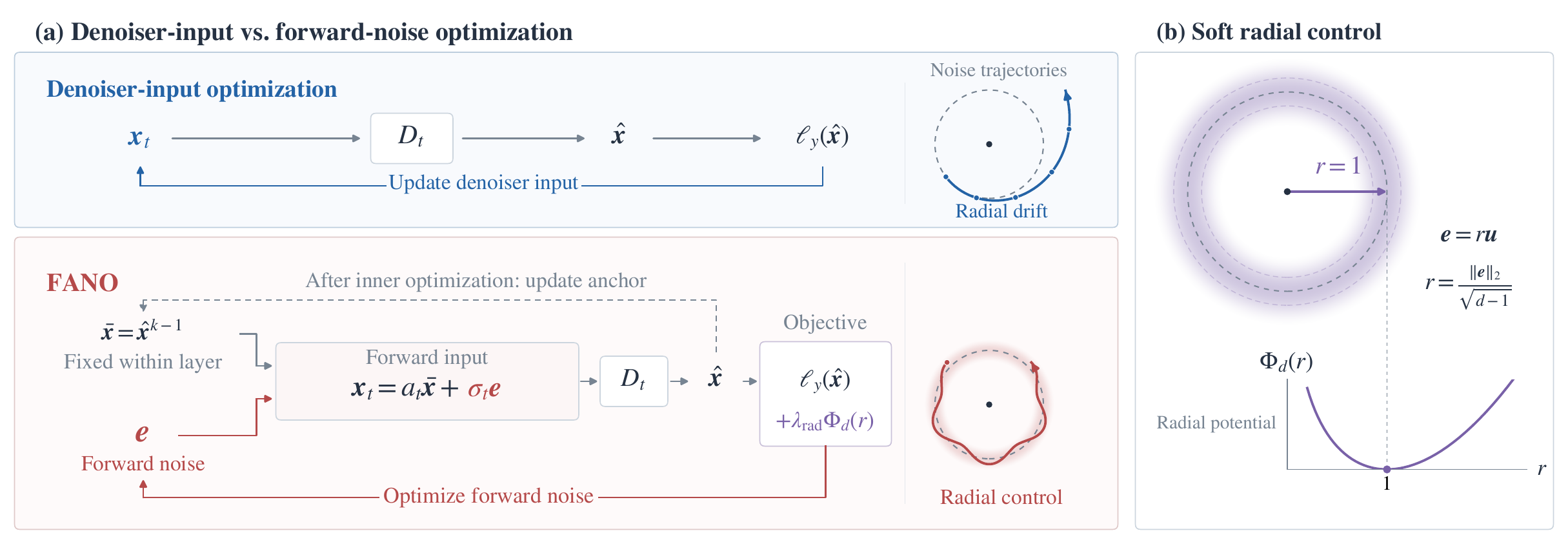}
\caption{Overview of FANO. \textbf{(a)}~Blue: direct optimization of $\vx_t$; red: FANO fixes the anchor $\bar\vx{=}\hat\vx^{k-1}$ and optimizes only the forward noise $\ve$. Circular diagrams contrast radial drift (blue) with controlled search (red). \textbf{(b)}~The radial potential $\Phi_d(r)$ is minimized at $r{=}1$, softly penalizing deviations from the typical noise shell. FANO decomposes $\ve{=}r\vu$ and jointly optimizes direction $\vu$ and magnitude $r$.}
\label{fig:main_overview}
\end{figure*}

Based on this analysis, we propose \emph{Anchored Forward-Noise Optimization} (FANO). Radial drift arises because the measurement-consistency objective does not constrain the norm of the noise component when directly optimizing $\vx_t$. FANO replaces the denoiser input $\vx_t$ with the forward noise $\ve$ as the search variable: each layer fixes the previous layer's denoised estimate as an anchor, keeps the denoiser input within the conditional forward distribution $q_t(\cdot\mid\bar\vx)$, and progressively contracts the search region through annealing. High-dimensional Gaussian concentration indicates that $\|\ve\|$ should concentrate near $\sqrt{d{-}1}$, and deviations degrade denoiser performance (Figure~\ref{fig:diagnosis}(b)). We therefore decompose $\ve$ into a direction and a radial magnitude in polar coordinates: the measurement loss drives the directional search, while a concentration-inspired potential anchors the radial magnitude near its typical value, keeping the denoiser input in the high-density region of the forward distribution. Experiments on 5 linear and 3 nonlinear inverse problems on FFHQ and ImageNet show that FANO matches the strongest current baselines on linear tasks and achieves consistent improvements on nonlinear tasks (Figure~\ref{fig:teaser}, Table~\ref{tab:main_results_005}). Figure~\ref{fig:main_overview} provides an overview of the method.

Our contributions are:
\begin{itemize}[nosep]
    \item \textbf{Diagnosis of radial drift.} We identify \emph{radial drift} in annealed methods that directly optimize $\vx_t$: the equivalent noise norm deviates from its typical value in later layers, degrading denoiser performance. Controlled experiments confirm the association and show that radial correction can recover the quality lost at the final layer.
    \item \textbf{Forward-noise optimization with radial control.} We propose FANO, which uses the forward noise $\ve$ as the within-layer search variable. This fixes the denoiser input to the forward-diffusion form and excludes the anchor from optimization. Exploiting the Gaussian prior, we decompose $\ve$ into a direction and a radial magnitude, with a concentration-inspired potential to keep the denoiser input in the high-density region.
    \item \textbf{Comprehensive evaluation.} Experiments on 5 linear and 3 nonlinear inverse problems show that FANO matches the strongest baselines on linear tasks and achieves consistent improvements on nonlinear tasks. Ablation studies identify the $\ve$-search framework as the primary source of gains.
\end{itemize}

\section{Background and Related Work}
\label{sec:background}

For zero-shot inverse problem solving, the denoiser provided by a pretrained diffusion model is directly used as an implicit image prior, incorporating measurement consistency into reverse sampling at inference time without retraining for each degradation type. Existing methods differ primarily in their search strategies and have broadly evolved along three directions: stepwise guidance locally corrects the diffusion state at each reverse step; end-to-end noise optimization encodes the reverse process as a deterministic mapping and performs a global search in the initial noise space; and annealed decoupling decomposes inference into a sequence of subproblems across noise levels, allowing a broader search within each layer.

\subsection{Diffusion Models}
\label{subsec:bg_diffusion_denoiser}

Diffusion models transform the data distribution $p(\vx_0)$ into an approximately Gaussian distribution through a forward process that progressively adds noise, and then learn a reverse denoising process to generate data from noise \citep{ho2020denoising,song2021scorebased}. Given noise-schedule parameters $(a_t,\sigma_t)$, the conditional distribution of the forward process at timestep $t$ is
\begin{equation}
    q_t(\vx_t\mid \vx_0)=\mathcal N(a_t \vx_0,\sigma_t^2 \mI_d),
    \qquad
    \vx_t=a_t \vx_0+\sigma_t\veps,
    \quad \veps\sim\mathcal N(\bm{0},\mI_d),
    \label{eq:bg_forward_marginal}
\end{equation}
Reverse generation is driven by the score function $\nabla_{\vx_t}\log p_t(\vx_t)$ \citep{anderson1982reverse,song2021denoising}, approximated in practice by a learned network $\bm{s}_\theta(\vx_t,t)$. Tweedie's formula \citep{boys2024tweedie} relates the score to $\mathbb E[\vx_0\mid \vx_t]$; substituting $\bm{s}_\theta$ defines the denoiser
\begin{equation}
    D_t(\vx_t):=\frac{1}{a_t}\bigl(\vx_t+\sigma_t^2 \bm{s}_\theta(\vx_t,t)\bigr)\approx \mathbb E[\vx_0\mid \vx_t],
    \label{eq:bg_denoiser}
\end{equation}
which provides the minimum mean-square-error estimate of $\vx_0$ given $\vx_t$.

\subsection{Stepwise Measurement Guidance}
\label{subsec:rw_stepwise_guidance}

DPS \citep{chung2023dps} approximates $p_t(\vy\mid \vx_t)$ via the denoised estimate $D_t(\vx_t)$, enabling guidance for arbitrary forward operators; subsequent work refines the approximation or exploits problem structure \citep{chung2022mcg,song2023pigdm,meng2022dmps}. Spectral-domain methods such as DDRM \citep{kawar2022ddrm} and DDNM \citep{wang2023ddnm} achieve exact data consistency on decomposable linear operators but do not extend to nonlinear problems. DSG \citep{yang2024dsg} and DiffRGD \citep{liao2026diffrgd} exploit the spherical geometry of high-dimensional Gaussians to mitigate guidance-induced distribution shift, the latter performing Riemannian optimization on a sphere of a given radius. Nonetheless, the correction range of stepwise guidance remains restricted to a neighborhood of the current diffusion state, making it difficult to traverse multiple posterior modes in nonlinear problems\citep{zhang2024daps}.

\subsection{End-to-End Noise Optimization}
\label{subsec:rw_e2e_noise_optimization}

End-to-end noise optimization treats the reverse process as a deterministic mapping $\mathcal{G}$ and directly solves $\min_{\vz}\|A(\mathcal{G}(\vz))-\vy\|^2$ in the initial noise space \citep{wang2024dmplug}. INO \citep{jia2026weak} constrains the search to the Gaussian typical shell $\|\vz\|=\sqrt{d}$ via tangent-space optimization; N-HMC \citep{xia2026nhmc} instead samples the posterior $p(\vz\mid \vy)$ with Hamiltonian Monte Carlo. This family avoids the $p_t(\vy\mid\vx_t)$ approximation but requires backpropagation through the entire generation chain, inherits the nonconvexity of the multistep composition, and loses the coarse-to-fine structure of diffusion.

\subsection{Annealed Decoupling}
\label{subsec:rw_annealing_methods}

Annealed decoupling methods decompose the reverse process into a sequence of within-layer subproblems along an annealing schedule, retaining diffusion's coarse-to-fine hierarchy\citep{rissanen2022generative,dieleman2024spectral} while allowing a broader search within each layer. Between layers, resampling---adding forward noise conditioned on a clean estimate \citep{lugmayr2022repaint}---returns the state to the forward distribution.

DAPS \citep{zhang2024daps} uses $\vx_0$ as the search variable and samples the within-layer posterior via Langevin dynamics; FAST-DIPS \citep{fastdips2026} replaces the Langevin chain with ADMM splitting for faster convergence \citep{li2024dcdp,zhan2026maprps}. SITCOM \citep{alkhouri2025sitcom} shifts the search variable to the denoiser input $\vx_t$, optimizing
$$\min_{\bm{v}_t}\|A(D_t(\bm{v}_t))-\vy\|^2+\lambda\|\vx_t-\bm{v}_t\|^2$$
and leveraging the denoiser's implicit bias for regularization. However, its Euclidean regularizer does not directly constrain the distributional structure of $\bm{v}_t$; within-layer optimization can drive $\bm{v}_t$ away from the high-density region of the forward distribution (Section~\ref{subsec:diagnosis}). FANO addresses this by using the forward noise $\ve$ as the search variable and introducing a concentration-inspired radial potential (Section~\ref{sec:method}).

\section{Method}
\label{sec:method}

This section first analyzes radial drift in existing annealed methods (Section~\ref{subsec:diagnosis}), then introduces an optimization strategy using the forward noise $\ve$ as the search variable together with a radial control mechanism (Section~\ref{subsec:method_shell}), and finally presents the complete algorithm, \textbf{FANO} (Anchored Forward-Noise Optimization, Section~\ref{subsec:method_algorithm}).

\begin{figure*}[t]
\centering
\includegraphics[width=\textwidth]{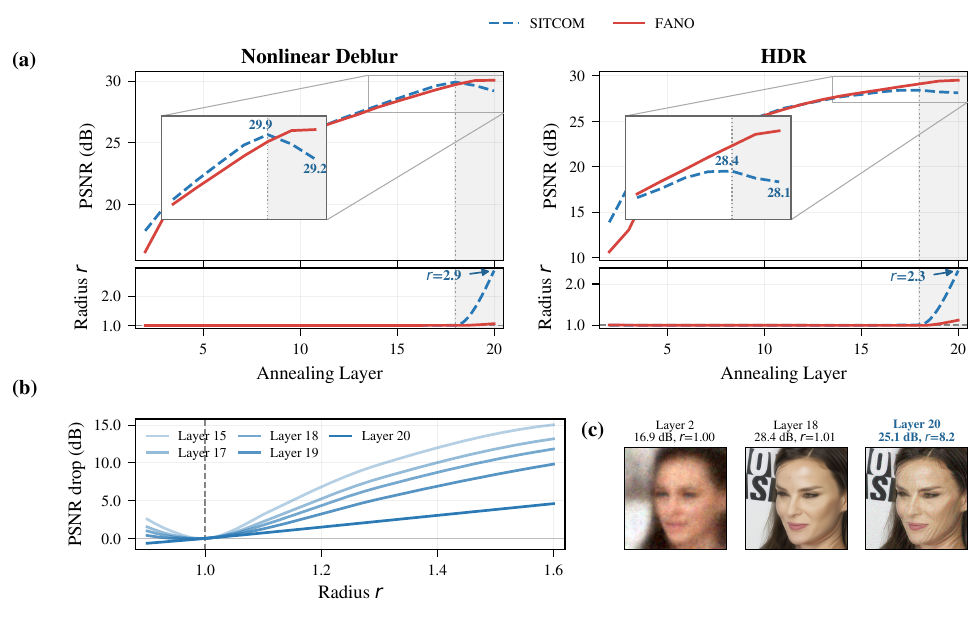}
\caption{Radial drift on FFHQ ($\sigma_y{=}0.05$, $N{=}20$). \textbf{(a)}~Mean PSNR and normalized radius $r$ on FFHQ100 for SITCOM \citep{alkhouri2025sitcom} and FANO; insets magnify late-layer PSNR, and shading marks layers after SITCOM's peak. \textbf{(b)}~Denoiser PSNR drop relative to $r{=}1$ when varying the radius along fixed random noise directions. \textbf{(c)}~SITCOM reconstructions at layers 2, 18, and 20, illustrating late-layer degradation as $r$ increases.}
\label{fig:diagnosis}
\end{figure*}

\subsection{Radial Drift in Annealed Decoupling}
\label{subsec:diagnosis}

Annealed decoupling methods share the following core recursion: given an annealing schedule $t_1>\cdots>t_N$, each layer adds fresh noise to the previous layer's denoised result and searches for a new estimate consistent with the measurement at the current noise level. Taking SITCOM \citep{alkhouri2025sitcom} as a representative method, we track this recursion (Figure~\ref{fig:diagnosis}(a)) and observe a puzzling phenomenon: on two nonlinear inverse problems, the measurement residual RMSE decreases monotonically throughout, and reconstruction PSNR improves over the first $18$ layers, yet abruptly reverses and decreases in the final two layers. The optimizer thus degrades reconstruction quality while successfully matching the measurement more closely. A natural alternative explanation is overfitting to measurement noise, but a control experiment with $\sigma_y=0$ rules out this hypothesis: without noise, the final-layer PSNR loss increases from $0.7$~dB to $3.4$~dB for nonlinear deblurring and from $0.3$~dB to $0.5$~dB for HDR. This more severe degradation indicates that the issue arises from the optimization dynamics themselves rather than noise overfitting.

To trace the cause of this degradation, we examine the variable optimized within each SITCOM layer. SITCOM directly optimizes the denoiser input $\vx_t$ at each annealing layer to make the denoised output match the measurement as closely as possible. Given the current layer's anchor $\bar\vx:=\hat\vx^{k-1}$, the forward process $\vx_t=a_t\bar\vx+\sigma_t\ve$ allows any denoiser input to be parameterized by an equivalent forward noise $\ve$. Define the normalized equivalent radius as $r:=\|\ve\|_2/\sqrt{d-1}$; under $\ve\sim\mathcal N(\bm{0},\mI_d)$, high-dimensional concentration ensures $r\approx1$. Tracking $r$ across layers (Figure~\ref{fig:diagnosis}(a), bottom row) reveals an anomalous signal coinciding with the degradation: over the first $18$ layers, $r\approx1.00$, but its mean rises to $2.9$ for nonlinear deblurring and $2.3$ for HDR in the final layer, with both statistics computed over the full FFHQ100 dataset. Panel~(c) visualizes the layerwise reconstructions of a single nonlinear deblurring example: at layer $18$, PSNR reaches $28.4$~dB ($r=1.01$); in the final layer, $r$ increases to $8.2$ and PSNR drops to $25.1$~dB.

Figure~\ref{fig:diagnosis}(b) tests the association between radial deviation and performance degradation through a controlled experiment. Holding a randomly sampled noise direction $\vu\sim\mathrm{Unif}(\mathbb S_{\sqrt{d-1}}^{d-1})$ fixed and scaling only the radius $r$, we find that denoiser performance is optimal near $r=1$ at every annealing layer. This test uses random directions, whereas directions obtained through multiple measurement-driven optimization steps are no longer random samples. We therefore further examine the terminal state of the final layer in the same SITCOM run, fixing the anchor and optimized direction but reconstructing the denoiser input as $a_t\bar\vx+\sigma_t(\ve/r)$ to normalize the equivalent noise's radial magnitude to $1$. This increases mean PSNR by $0.15$~dB for HDR and $0.49$~dB for nonlinear deblurring, recovering most of the final-layer degradation by correcting the radius alone. We refer to this late-stage degradation caused by the radial magnitude deviating from its typical value as \emph{radial drift}. Note that $r$ is measured relative to the estimated anchor rather than the ground truth, so anchor error can partially amplify the observed $r$. The red curves in Figure~\ref{fig:diagnosis}(a) show the layerwise behavior of our method, FANO (Sections~\ref{subsec:method_shell}--\ref{subsec:method_algorithm}), under the same conditions: mean PSNR continues to improve through the final layer, the normalized radius remains near $1$, and radial drift disappears.

\subsection{Forward-Noise Optimization and Radial Control}
\label{subsec:method_shell}

The preceding analysis directly links radial drift to the uncontrolled norm of the noise component during within-layer optimization. FANO therefore shifts the search variable from $\vx_t$ to the forward noise $\ve$: at each annealing layer, it fixes the previous layer's denoised estimate $\bar\vx:=\hat\vx^{k-1}$ as an anchor and parameterizes the denoiser input as $\vx_t=a_t\bar\vx+\sigma_t \ve$, using $\ve$ as the sole within-layer search variable. This parameterization preserves the forward-diffusion structure $q_t(\cdot\mid\bar\vx)$ of the denoiser input. Searching over $\ve$ treats the anchor $\bar\vx$ as a fixed offset excluded from within-layer optimization, so within-layer updates modify only the noise component.

The prior $\mathcal N(\bm{0},\mI_d)$ is uniform in direction and highly concentrated in radius (Figure~\ref{fig:diagnosis}(b)), motivating a polar decomposition of $\ve$ into a radial magnitude $r$ and a direction $\vu$. The measurement loss drives the directional search, while a concentration-inspired soft constraint controls the radius.

We decompose the forward noise $\ve\in\mathbb R^d$ into a normalized radial magnitude and a direction as $\ve=r\cdot \vu$, where $r:=\|\ve\|_2/\sqrt{d-1}>0$. Let $\mathbb S:=\mathbb S_{\sqrt{d-1}}^{d-1}:=\{\vu\in\mathbb R^d:\|\vu\|_2=\sqrt{d-1}\}$, so that $\vu\in\mathbb S$. Under the polar change of variables, the joint density of the standard Gaussian prior $\mathcal N(\bm{0},\mI_d)$ factorizes as
\begin{equation}
    p(r,\vu)\propto r^{d-1}\exp\!\left(-\frac{(d-1)\,r^2}{2}\right),
    \label{eq:polar_prior}
\end{equation}
where the direction $\vu$ is uniformly distributed on $\mathbb S$ and the radius $r$ follows a normalized $\chi$ distribution\citep{laurent2000adaptive}. The prior has no directional preference but strongly favors a particular radial range, leading to the following constraint strategy: \emph{constrain only the radial dimension and let the measurement loss fully drive the angular search}.

The negative log of the radial prior yields the radial potential
\begin{equation}
    \Phi_d(r)=\frac{d-1}{2}\bigl(r^2-1-2\log r\bigr),
    \label{eq:radial_potential}
\end{equation}
which attains its minimum value of zero at $r=1$ (i.e., $\|\ve\|_2=\sqrt{d-1}$; see Appendix~\ref{app:radial_potential_derivation} for the derivation), providing soft radial control of the denoiser input inspired by Gaussian radial statistics. Let $\ell_y(\vx):=\|A(\vx)-\vy\|_2^2$ denote the measurement loss. For a fixed annealing layer $t$ and anchor $\bar\vx$, the measurement loss along direction $\vu$ is $F_t(r;\vu):=\ell_y(D_t(a_t\bar\vx+\sigma_t r\vu))$. Replacing the coefficient prescribed by the prior with a tunable hyperparameter $\lambda_{\mathrm{rad}}\ge0$ gives FANO's within-layer objective:
\begin{equation}
    \begin{aligned}
        \min_{r,\vu}\quad
        J_\lambda(r;\vu)&:=F_t(r;\vu)+\lambda_{\mathrm{rad}}\,\Phi_d(r)\\
        \text{s.t.}\quad
        &\vu\in\mathbb S,\quad r>0.
    \end{aligned}
    \label{eq:polar_map}
\end{equation}
The following proposition characterizes the optimization landscape of the one-dimensional radial subproblem $\min_{r>0} J_\lambda(r;\vu)$.

\begin{proposition}[Boundedness and benefits of radial relaxation]
\label{prop:radial_relaxation}
Assume $d\ge2$, $\lambda_{\mathrm{rad}}>0$, and that $F_t(\cdot\,;\vu)$ is nonnegative and continuous on $(0,\infty)$. Then, $\forall\,\vu\in\mathbb S$:
\begin{enumerate}[label=(\roman*),nosep,leftmargin=*]
\item $J_\lambda(\cdot\,;\vu)$ attains a global minimum over $r>0$ at $r_\lambda(\vu)$, and
\begin{equation}
  |r_\lambda(\vu)-1|
  \;\le\;
  \sqrt{\frac{2\,F_t(1;\vu)}{\lambda_{\mathrm{rad}}(d-1)}}\,.
  \label{eq:radial_bound}
\end{equation}
\item If $F_t(\cdot\,;\vu)$ is differentiable at $r{=}1$ and $\partial_r F_t(1;\vu)\neq 0$, then
  $J_\lambda(r_\lambda(\vu);\vu)<J_\lambda(1;\vu)=F_t(1;\vu)$.
\end{enumerate}
\end{proposition}

Part (i) guarantees that the optimal radius does not diverge: the radial potential anchors $r_\lambda$ near $1$, with the deviation shrinking as $\lambda_{\mathrm{rad}}$ increases. Part (ii) shows that when the measurement loss has a nonzero radial gradient, hard spherical search with $r\equiv 1$ cannot exploit this descent direction, whereas the soft constraint permits a bounded deviation and strictly reduces the objective. This result is consistent with the ablation in Table~\ref{tab:ablation}, where FANO outperforms FANO-Sphere on HDR by approximately $0.8$\,dB. The proof is provided in Appendix~\ref{app:radial_relaxation_proof}, and the strong-regularization asymptotic expansion in Corollary~\ref{cor:asymptotic}.

\subsection{Algorithm}
\label{subsec:method_algorithm}

FANO proceeds recursively along an annealing schedule $t_1>t_2>\dots>t_N$. Layer $k$ uses the denoised estimate from layer $k-1$, $\hat\vx^{k-1}$, as its anchor. Each layer performs three steps: \emph{(i) Initialization}: sample a random direction $\vu$ on the sphere and set the radius to $r=1$; \emph{(ii) Forward-noise optimization}: solve Equation~\eqref{eq:polar_map} for $r^\star,\vu^\star$, using Riemannian Adam for the direction $\vu$ \citep{absil2008optimization} and standard Adam for the radius $r$. Each inner step requires only one forward and one backward pass through the denoiser to obtain gradients for both variables, requiring, over $N$ layers with at most $K$ steps per layer, at most $NK$ NFEs; \emph{(iii) Anchor update}: pass the optimized denoised estimate to the next layer:
\begin{equation}
    \hat\vx^{k}:=D_{t_k}\bigl(a_{t_k}\hat\vx^{k-1}+\sigma_{t_k}r^\star \vu^\star\bigr).
    \label{eq:anchor_recursion}
\end{equation}
The initial anchor $\hat\vx^0$ is obtained by denoising pure noise with $D_{t_1}$.

Within-layer iterations use the measurement residual for early stopping. The current layer terminates when
\begin{equation}
    \bigl\|A\bigl(D_{t_k}(\vz_j)\bigr)-\vy\bigr\|_2\ \le\ \tau_{\mathrm{stop}}:=\sigma_y\sqrt m
    \label{eq:gaussian_tau_stop}
\end{equation}
where $\sigma_y$ is the standard deviation of the measurement noise and $m$ is the measurement dimension. The quantity $\sigma_y\sqrt m$ is the typical norm of additive Gaussian noise $\vxi\sim\mathcal N(\bm{0},\sigma_y^2\mI_m)$; a residual below this level indicates that optimization is beginning to fit the noise itself. In the noiseless case, $\tau_{\mathrm{stop}}=0$ and the criterion reduces to a fixed iteration budget. The full pseudocode is provided in Appendix~\ref{app:algorithm}, and optimizer implementation details in Appendix~\ref{app:optimizer_details}.

\section{Experiments}
\label{sec:experiments}

We evaluate FANO on 5 linear tasks (super-resolution, Gaussian deblurring, motion deblurring, box inpainting, and random inpainting) and 3 nonlinear tasks (phase retrieval, high dynamic range (HDR) reconstruction, and nonlinear deblurring).

\subsection{Experimental Setup}
\label{subsec:exp_setup}

\paragraph{Datasets, tasks, and metrics}
Experiments are conducted on FFHQ $256\times 256$ \citep{karras2019style} and ImageNet $256\times 256$ \citep{deng2009imagenet}, using 100 test images from each dataset. We use publicly available pretrained VP SDE DDPMs: the FFHQ model is from \cite{choi2021ilvr}, and the ImageNet model is from \cite{dhariwal2021diffusion}, without task-specific fine-tuning. Degradation operators for linear tasks follow the standard settings of DAPS \citep{zhang2024daps}. For nonlinear tasks, phase retrieval uses an oversampling ratio of $1.5$ (a resolution ratio of $384/256$, corresponding to a $384\times 384$ FFT; DAPS \citep{zhang2024daps} denotes this as OS${}=2$), HDR reconstruction uses a clipping factor of $2$, and the spatially varying blur kernels for nonlinear deblurring are from BKSE \citep{tran2021explore}. The measurement noise level is $\sigma_y=0.05$ in our experiments; task-specific degradation parameters are detailed in Appendix~\ref{app:task_details}. We report PSNR$\uparrow$, SSIM$\uparrow$ \citep{wang2004image}, and LPIPS$\downarrow$ \citep{zhang2018unreasonable}, with LPIPS consistently computed using a VGG16 feature network.

\paragraph{Methods and configurations}
We compare against DAPS \citep{zhang2024daps}, SITCOM \citep{alkhouri2025sitcom}, FAST-DIPS \citep{fastdips2026}, DiffRGD \citep{liao2026diffrgd}, and INO \citep{jia2026weak}. All methods are rerun using their public code and recommended hyperparameters with the same checkpoints, test images, degradation operators, and noise levels. Table~\ref{tab:main_results_005} reports each baseline on the tasks supported by its public code: DiffRGD covers linear tasks except box inpainting, and INO does not cover motion deblurring, phase retrieval, or HDR. FANO uses the same algorithmic framework across all tasks (Appendix Algorithm~\ref{alg:fano}), with learning rates $\eta_u=\eta_r=0.02$ and the noise-ball early-stopping criterion in Equation~\eqref{eq:gaussian_tau_stop}; detailed configurations are listed in Appendix~\ref{app:hyperparameters}. Following the convention in prior work \citep{chung2023dps,zhang2024daps,alkhouri2025sitcom}, phase retrieval uses 4 independent runs per image and reports the best result; all baselines use the same protocol.

\begin{table*}[t!]
\centering
\caption{Quantitative results at $\sigma_y=0.05$, reported as the mean $\pm$ standard deviation over 100 test images. The best results are in \textbf{bold}, and the second-best results are \underline{underlined}.}
\label{tab:main_results_005}
\footnotesize
\setlength{\tabcolsep}{3.4pt}
\renewcommand{\arraystretch}{1.0}
\begin{tabular}{ll ccc ccc}
\toprule
& & \multicolumn{3}{c}{\textbf{FFHQ}} & \multicolumn{3}{c}{\textbf{ImageNet}} \\
\cmidrule(lr){3-5} \cmidrule(lr){6-8}
\textbf{Task} & \textbf{Method} & PSNR$\uparrow$ & SSIM$\uparrow$ & LPIPS$\downarrow$ & PSNR$\uparrow$ & SSIM$\uparrow$ & LPIPS$\downarrow$ \\
\midrule
\multirow{6}{*}{SR $\times 4$}
& DAPS     & 29.35{\scriptsize$\pm$1.47} & 0.782{\scriptsize$\pm$0.028} & 0.268{\scriptsize$\pm$0.035} & 25.47{\scriptsize$\pm$2.86} & 0.638{\scriptsize$\pm$0.087} & 0.376{\scriptsize$\pm$0.061} \\
& SITCOM   & \underline{30.55{\scriptsize$\pm$1.87}} & \underline{0.868{\scriptsize$\pm$0.033}} & \underline{0.223{\scriptsize$\pm$0.034}} & \textbf{27.09{\scriptsize$\pm$3.84}} & \textbf{0.757{\scriptsize$\pm$0.116}} & \textbf{0.307{\scriptsize$\pm$0.077}} \\
& FAST-DIPS & 29.59{\scriptsize$\pm$1.75} & 0.843{\scriptsize$\pm$0.041} & 0.241{\scriptsize$\pm$0.042} & 26.55{\scriptsize$\pm$3.65} & 0.725{\scriptsize$\pm$0.125} & 0.325{\scriptsize$\pm$0.072} \\
& DiffRGD  & 29.24{\scriptsize$\pm$1.93} & 0.838{\scriptsize$\pm$0.042} & \textbf{0.214{\scriptsize$\pm$0.037}} & 25.96{\scriptsize$\pm$3.70} & 0.712{\scriptsize$\pm$0.136} & 0.322{\scriptsize$\pm$0.084} \\
& INO      & 29.24{\scriptsize$\pm$1.80} & 0.809{\scriptsize$\pm$0.037} & 0.260{\scriptsize$\pm$0.038} & 25.65{\scriptsize$\pm$3.50} & 0.694{\scriptsize$\pm$0.131} & \underline{0.310{\scriptsize$\pm$0.090}} \\
& \textbf{FANO}    & \textbf{30.68{\scriptsize$\pm$1.88}} & \textbf{0.872{\scriptsize$\pm$0.032}} & 0.232{\scriptsize$\pm$0.036} & \underline{27.03{\scriptsize$\pm$3.87}} & \underline{0.755{\scriptsize$\pm$0.118}} & 0.321{\scriptsize$\pm$0.082} \\
\midrule
\multirow{6}{*}{Gauss.\ deblur}
& DAPS     & 29.61{\scriptsize$\pm$1.63} & 0.786{\scriptsize$\pm$0.031} & 0.259{\scriptsize$\pm$0.032} & 25.88{\scriptsize$\pm$3.23} & 0.657{\scriptsize$\pm$0.105} & 0.356{\scriptsize$\pm$0.062} \\
& SITCOM   & \underline{29.89{\scriptsize$\pm$1.95}} & \underline{0.849{\scriptsize$\pm$0.038}} & \underline{0.242{\scriptsize$\pm$0.038}} & \underline{26.38{\scriptsize$\pm$3.60}} & \underline{0.726{\scriptsize$\pm$0.128}} & 0.340{\scriptsize$\pm$0.083} \\
& FAST-DIPS & 29.20{\scriptsize$\pm$1.92} & 0.830{\scriptsize$\pm$0.045} & 0.259{\scriptsize$\pm$0.045} & 25.92{\scriptsize$\pm$3.74} & 0.694{\scriptsize$\pm$0.142} & 0.358{\scriptsize$\pm$0.082} \\
& DiffRGD  & 27.05{\scriptsize$\pm$5.12} & 0.780{\scriptsize$\pm$0.162} & 0.248{\scriptsize$\pm$0.106} & 22.87{\scriptsize$\pm$3.36} & 0.564{\scriptsize$\pm$0.171} & 0.413{\scriptsize$\pm$0.091} \\
& INO      & 28.72{\scriptsize$\pm$1.88} & 0.789{\scriptsize$\pm$0.048} & 0.275{\scriptsize$\pm$0.042} & 24.76{\scriptsize$\pm$3.46} & 0.653{\scriptsize$\pm$0.145} & \underline{0.334{\scriptsize$\pm$0.087}} \\
& \textbf{FANO}    & \textbf{30.65{\scriptsize$\pm$2.04}} & \textbf{0.868{\scriptsize$\pm$0.036}} & \textbf{0.225{\scriptsize$\pm$0.038}} & \textbf{27.01{\scriptsize$\pm$3.95}} & \textbf{0.746{\scriptsize$\pm$0.128}} & \textbf{0.317{\scriptsize$\pm$0.084}} \\
\midrule
\multirow{5}{*}{Motion deblur}
& DAPS     & 31.65{\scriptsize$\pm$1.34} & 0.836{\scriptsize$\pm$0.022} & 0.208{\scriptsize$\pm$0.029} & 28.65{\scriptsize$\pm$3.11} & 0.769{\scriptsize$\pm$0.066} & 0.244{\scriptsize$\pm$0.055} \\
& SITCOM   & \underline{32.37{\scriptsize$\pm$1.70}} & \underline{0.897{\scriptsize$\pm$0.024}} & \underline{0.188{\scriptsize$\pm$0.031}} & \underline{29.38{\scriptsize$\pm$3.68}} & \underline{0.832{\scriptsize$\pm$0.080}} & \underline{0.235{\scriptsize$\pm$0.062}} \\
& FAST-DIPS & 31.38{\scriptsize$\pm$1.59} & 0.873{\scriptsize$\pm$0.035} & 0.200{\scriptsize$\pm$0.036} & 28.79{\scriptsize$\pm$3.92} & 0.796{\scriptsize$\pm$0.106} & 0.261{\scriptsize$\pm$0.078} \\
& DiffRGD  & 29.24{\scriptsize$\pm$2.07} & 0.835{\scriptsize$\pm$0.044} & 0.213{\scriptsize$\pm$0.035} & 23.40{\scriptsize$\pm$3.44} & 0.591{\scriptsize$\pm$0.169} & 0.421{\scriptsize$\pm$0.100} \\
& \textbf{FANO}    & \textbf{33.04{\scriptsize$\pm$1.82}} & \textbf{0.909{\scriptsize$\pm$0.024}} & \textbf{0.167{\scriptsize$\pm$0.029}} & \textbf{29.88{\scriptsize$\pm$4.19}} & \textbf{0.840{\scriptsize$\pm$0.085}} & \textbf{0.225{\scriptsize$\pm$0.070}} \\
\midrule
\multirow{5}{*}{Inpaint (box)}
& DAPS     & 24.32{\scriptsize$\pm$2.67} & 0.747{\scriptsize$\pm$0.031} & 0.236{\scriptsize$\pm$0.034} & \underline{21.05{\scriptsize$\pm$3.44}} & 0.713{\scriptsize$\pm$0.036} & 0.271{\scriptsize$\pm$0.057} \\
& SITCOM   & 24.47{\scriptsize$\pm$2.52} & 0.737{\scriptsize$\pm$0.033} & 0.245{\scriptsize$\pm$0.030} & 19.62{\scriptsize$\pm$2.91} & 0.699{\scriptsize$\pm$0.038} & \underline{0.255{\scriptsize$\pm$0.060}} \\
& FAST-DIPS & \underline{25.18{\scriptsize$\pm$2.45}} & \underline{0.857{\scriptsize$\pm$0.028}} & \underline{0.182{\scriptsize$\pm$0.030}} & \textbf{21.53{\scriptsize$\pm$4.17}} & \textbf{0.797{\scriptsize$\pm$0.058}} & 0.257{\scriptsize$\pm$0.044} \\
& INO      & 22.79{\scriptsize$\pm$2.91} & 0.827{\scriptsize$\pm$0.039} & 0.222{\scriptsize$\pm$0.041} & 18.69{\scriptsize$\pm$3.02} & 0.739{\scriptsize$\pm$0.105} & 0.282{\scriptsize$\pm$0.070} \\
& \textbf{FANO}    & \textbf{25.33{\scriptsize$\pm$2.94}} & \textbf{0.866{\scriptsize$\pm$0.028}} & \textbf{0.156{\scriptsize$\pm$0.030}} & 20.40{\scriptsize$\pm$3.30} & \underline{0.786{\scriptsize$\pm$0.040}} & \textbf{0.206{\scriptsize$\pm$0.040}} \\
\midrule
\multirow{6}{*}{Inpaint (rand.)}
& DAPS     & 30.72{\scriptsize$\pm$1.35} & 0.801{\scriptsize$\pm$0.020} & 0.230{\scriptsize$\pm$0.026} & 27.35{\scriptsize$\pm$3.20} & 0.725{\scriptsize$\pm$0.066} & 0.248{\scriptsize$\pm$0.050} \\
& SITCOM   & 32.40{\scriptsize$\pm$1.44} & 0.863{\scriptsize$\pm$0.018} & 0.182{\scriptsize$\pm$0.030} & 28.90{\scriptsize$\pm$3.57} & 0.814{\scriptsize$\pm$0.054} & \underline{0.180{\scriptsize$\pm$0.053}} \\
& FAST-DIPS & 31.78{\scriptsize$\pm$1.57} & 0.892{\scriptsize$\pm$0.030} & 0.185{\scriptsize$\pm$0.034} & \underline{29.46{\scriptsize$\pm$4.11}} & \underline{0.828{\scriptsize$\pm$0.093}} & 0.214{\scriptsize$\pm$0.058} \\
& DiffRGD  & 32.73{\scriptsize$\pm$1.91} & 0.912{\scriptsize$\pm$0.023} & 0.167{\scriptsize$\pm$0.030} & 28.90{\scriptsize$\pm$4.15} & 0.825{\scriptsize$\pm$0.094} & 0.239{\scriptsize$\pm$0.070} \\
& INO      & \underline{32.97{\scriptsize$\pm$2.14}} & \underline{0.915{\scriptsize$\pm$0.031}} & \underline{0.166{\scriptsize$\pm$0.035}} & 28.33{\scriptsize$\pm$4.55} & 0.810{\scriptsize$\pm$0.124} & 0.218{\scriptsize$\pm$0.082} \\
& \textbf{FANO}    & \textbf{34.07{\scriptsize$\pm$1.97}} & \textbf{0.931{\scriptsize$\pm$0.017}} & \textbf{0.120{\scriptsize$\pm$0.027}} & \textbf{29.94{\scriptsize$\pm$4.61}} & \textbf{0.864{\scriptsize$\pm$0.074}} & \textbf{0.157{\scriptsize$\pm$0.050}} \\
\midrule
\multirow{4}{*}{Phase retrieval}
& DAPS     & \underline{30.09{\scriptsize$\pm$2.96}} & 0.781{\scriptsize$\pm$0.057} & 0.239{\scriptsize$\pm$0.058} & \underline{22.94{\scriptsize$\pm$6.22}} & 0.505{\scriptsize$\pm$0.216} & \underline{0.411{\scriptsize$\pm$0.148}} \\
& SITCOM   & 29.21{\scriptsize$\pm$4.99} & 0.817{\scriptsize$\pm$0.178} & 0.264{\scriptsize$\pm$0.122} & 20.49{\scriptsize$\pm$6.38} & 0.467{\scriptsize$\pm$0.274} & 0.465{\scriptsize$\pm$0.163} \\
& FAST-DIPS & 28.37{\scriptsize$\pm$3.71} & \underline{0.849{\scriptsize$\pm$0.094}} & \underline{0.221{\scriptsize$\pm$0.085}} & 21.38{\scriptsize$\pm$6.72} & \underline{0.548{\scriptsize$\pm$0.274}} & 0.440{\scriptsize$\pm$0.191} \\
& \textbf{FANO}    & \textbf{31.98{\scriptsize$\pm$4.14}} & \textbf{0.877{\scriptsize$\pm$0.118}} & \textbf{0.187{\scriptsize$\pm$0.091}} & \textbf{23.83{\scriptsize$\pm$7.99}} & \textbf{0.567{\scriptsize$\pm$0.289}} & \textbf{0.388{\scriptsize$\pm$0.187}} \\
\midrule
\multirow{4}{*}{HDR}
& DAPS     & 27.48{\scriptsize$\pm$3.64} & \underline{0.837{\scriptsize$\pm$0.082}} & \underline{0.210{\scriptsize$\pm$0.073}} & \underline{26.07{\scriptsize$\pm$4.08}} & \underline{0.828{\scriptsize$\pm$0.096}} & \underline{0.204{\scriptsize$\pm$0.110}} \\
& SITCOM   & \underline{28.13{\scriptsize$\pm$3.12}} & 0.810{\scriptsize$\pm$0.107} & 0.217{\scriptsize$\pm$0.086} & 25.75{\scriptsize$\pm$3.66} & 0.771{\scriptsize$\pm$0.147} & 0.232{\scriptsize$\pm$0.131} \\
& FAST-DIPS & 25.90{\scriptsize$\pm$3.25} & \underline{0.837{\scriptsize$\pm$0.077}} & 0.227{\scriptsize$\pm$0.071} & 24.58{\scriptsize$\pm$4.04} & 0.809{\scriptsize$\pm$0.113} & 0.237{\scriptsize$\pm$0.116} \\
& \textbf{FANO}    & \textbf{29.88{\scriptsize$\pm$3.42}} & \textbf{0.878{\scriptsize$\pm$0.082}} & \textbf{0.157{\scriptsize$\pm$0.076}} & \textbf{27.32{\scriptsize$\pm$3.68}} & \textbf{0.832{\scriptsize$\pm$0.116}} & \textbf{0.196{\scriptsize$\pm$0.113}} \\
\midrule
\multirow{5}{*}{NL deblur}
& DAPS     & 28.53{\scriptsize$\pm$1.53} & 0.763{\scriptsize$\pm$0.032} & 0.259{\scriptsize$\pm$0.034} & 27.25{\scriptsize$\pm$3.13} & 0.720{\scriptsize$\pm$0.080} & 0.284{\scriptsize$\pm$0.064} \\
& SITCOM   & \underline{29.31{\scriptsize$\pm$2.17}} & 0.801{\scriptsize$\pm$0.085} & \underline{0.240{\scriptsize$\pm$0.073}} & \underline{28.47{\scriptsize$\pm$3.96}} & \underline{0.802{\scriptsize$\pm$0.094}} & \textbf{0.204{\scriptsize$\pm$0.064}} \\
& FAST-DIPS & 27.55{\scriptsize$\pm$1.63} & \underline{0.803{\scriptsize$\pm$0.044}} & 0.282{\scriptsize$\pm$0.048} & 25.50{\scriptsize$\pm$3.59} & 0.696{\scriptsize$\pm$0.140} & 0.383{\scriptsize$\pm$0.111} \\
& INO      & 27.53{\scriptsize$\pm$2.41} & 0.797{\scriptsize$\pm$0.056} & 0.252{\scriptsize$\pm$0.041} & 25.39{\scriptsize$\pm$3.74} & 0.688{\scriptsize$\pm$0.137} & 0.318{\scriptsize$\pm$0.081} \\
& \textbf{FANO}    & \textbf{30.83{\scriptsize$\pm$1.77}} & \textbf{0.873{\scriptsize$\pm$0.031}} & \textbf{0.183{\scriptsize$\pm$0.033}} & \textbf{29.09{\scriptsize$\pm$4.25}} & \textbf{0.819{\scriptsize$\pm$0.100}} & \underline{0.222{\scriptsize$\pm$0.072}} \\
\bottomrule
\end{tabular}
\end{table*}

\subsection{Quantitative Results}
\label{subsec:quantitative_results}

Table~\ref{tab:main_results_005} summarizes the quantitative results across all tasks. On linear inverse problems, FANO achieves reconstruction quality comparable to or slightly better than the strongest current baselines: it outperforms the second-best method in PSNR by 0.48--1.10~dB on random inpainting, 0.63--0.76~dB on Gaussian deblurring, and 0.50--0.67~dB on motion deblurring, with box inpainting remaining a relative weakness. The performance gap is more pronounced on nonlinear tasks: on FFHQ, FANO exceeds the second-best method in PSNR by 1.89~dB for phase retrieval, 1.52~dB for nonlinear deblurring, and 1.75~dB for HDR reconstruction. Its advantage on nonlinear tasks is particularly pronounced relative to SITCOM, which shares the annealing framework but searches in Euclidean space. This is consistent with the analysis in Section~\ref{subsec:diagnosis}: radial drift occurs more frequently with nonlinear measurement operators, and soft radial control effectively keeps the denoiser input within its operating regime.

Appendix~\ref{app:budget_analysis} compares the computational costs and runtimes of the main methods. FANO's reconstruction quality degrades smoothly as measurement noise increases, with detailed noise-robustness comparisons provided in Appendix~\ref{app:noise_robustness}. Figure~\ref{fig:teaser} presents visual reconstruction comparisons on representative tasks, and qualitative results for additional tasks are provided in Appendix~\ref{app:qualitative}. Appendix~\ref{app:pr_no_oversampling} further explores the ability of generative priors to compensate for missing oversampling information in phase retrieval without oversampling ($\text{OS}=1$).

\subsection{Ablation Studies}
\label{subsec:ablation}

\begin{table}[ht]
\centering
\caption{Ablation of the polar parameterization and radial freedom (FFHQ100, $\sigma_y=0.05$). The three variants share the $\ve$-search framework ($N$, $K$, and residual-based early stopping) and differ in two design choices: \emph{Polar} (the polar decomposition $\ve=r\vu$) and \emph{Radial freedom} (allowing $r$ to vary freely; otherwise, $r\equiv1$ is fixed). The best results are in \textbf{bold}.}
\label{tab:ablation}
\small
\setlength{\tabcolsep}{3.8pt}
\renewcommand{\arraystretch}{1.08}
\begin{tabular}{lcc ccc ccc}
\toprule
& \multicolumn{2}{c}{\textbf{Component}} & \multicolumn{3}{c}{\textbf{HDR}} & \multicolumn{3}{c}{\textbf{NL deblur}} \\
\cmidrule(lr){2-3} \cmidrule(lr){4-6} \cmidrule(lr){7-9}
\textbf{Variant} & \rotatebox{0}{\small Polar} & \rotatebox{0}{\small Radial freedom} & PSNR & SSIM & LPIPS & PSNR & SSIM & LPIPS \\
\midrule
FANO       & \checkmark & \checkmark & \textbf{29.88} & \textbf{0.878} & \textbf{0.157} & \textbf{30.83} & \textbf{0.873} & 0.183 \\
FANO-Sphere & \checkmark & $\times$  & 29.06 & 0.858 & 0.167 & 30.72 & 0.870 & 0.183 \\
FANO-Euc   & $\times$  & \checkmark & 29.18 & 0.859 & 0.171 & 30.80 & 0.871 & \textbf{0.182} \\
\bottomrule
\end{tabular}
\end{table}

Our ablation isolates the respective contributions of the polar parameterization and radial freedom (Table~\ref{tab:ablation}). All three variants share the $\ve$-search framework ($N$, $K$, residual-based early stopping, and hyperparameters), differing only in parameterization and radial freedom:
\begin{itemize}
    \item \textbf{FANO} (default): uses the polar decomposition $\ve=r\vu$ and jointly optimizes $(r,\vu)$;
    \item \textbf{FANO-Sphere}: uses the polar decomposition, fixes $r\equiv 1$, and optimizes only the direction $\vu$;
    \item \textbf{FANO-Euc}: directly optimizes $\ve$ with standard Adam, retaining the radial potential $\lambda_{\mathrm{rad}}\Phi_d(\|\ve\|_2/\sqrt{d{-}1})$ but without the polar decomposition.
\end{itemize}
All three variants outperform every comparison method in Table~\ref{tab:main_results_005} in PSNR, SSIM, and LPIPS on HDR and nonlinear deblurring, indicating that the $\ve$-search framework itself provides the primary gains. In particular, FANO-Sphere fixes $r\equiv1$ and requires no tuning of $\lambda_{\mathrm{rad}}$, yet its PSNR still exceeds the strongest comparison method by $0.93$~dB and $1.41$~dB on the two tasks, respectively, making it a ready-to-use, tuning-free default. The complete comparison across $8$ tasks is given in Appendix Table~\ref{tab:app_sphere_baseline}, and sensitivity to $\lambda_{\mathrm{rad}}$ is analyzed in Appendix~\ref{app:lambda_sensitivity}.

Differences within the framework are task-dependent. On nonlinear deblurring, the PSNR gap among the three variants is at most $0.11$~dB, with no notable separation in SSIM or LPIPS either. This indicates that the $\ve$-search framework already realizes its advantages on this task, leaving limited additional benefit from the polar decomposition or radial freedom. HDR behaves differently: allowing radial deviations improves FANO over FANO-Sphere by $0.82$~dB, while the polar decomposition improves FANO over FANO-Euc by $0.70$~dB. Both gains also hold for SSIM and LPIPS, indicating independent contributions from the two mechanisms.

\FloatBarrier
\section{Conclusion}
\label{sec:conclusion}

We analyze \emph{radial drift} in annealed diffusion methods for inverse problems: within-layer optimization drives the equivalent noise norm of the denoiser input away from the typical value of the Gaussian prior, degrading reconstruction quality in later layers. To address this limitation, we propose FANO, which uses the forward noise $\ve$ instead of the denoiser input $\vx_t$ as the within-layer search variable in an annealing framework. The anchor is excluded from within-layer updates, the denoiser input retains the forward-diffusion structure, and the search region contracts progressively through annealing. The Gaussian prior on $\ve$ further provides a statistical basis for radial control, with a concentration-inspired radial potential keeping the denoiser input in the high-density region of the forward distribution. Experiments on 8 linear and nonlinear tasks validate the effectiveness of FANO. Ablation studies identify the $\ve$-search framework as the primary source of improvement, and the gains on nonlinear tasks are consistent with the observations from our radial drift analysis.

The current framework relies on deterministic optimization and restricts the search space to a neighborhood of a single optimization trajectory, preventing exploration of the multimodal structure of the inverse problem posterior. Incorporating the geometric structure of noise into a stochastic sampling framework, for example by constructing Langevin dynamics on the sphere, offers a possible path from single-point reconstruction to sampling multiple solutions and quantifying uncertainty.

\subsection*{Reproducibility statement}

Algorithm~\ref{alg:fano} provides the complete pseudocode for FANO. All hyperparameters (annealing layers, inner steps, learning rates, and radial regularization coefficients) are listed in Appendix Table~\ref{tab:app_fano_params}; task-specific degradation parameters are given in Appendix Table~\ref{tab:app_task_params}; and the Riemannian Adam optimizer and positivity safeguards are detailed in Appendix~\ref{app:optimizer_details}. All experiments use publicly available pretrained diffusion models from \citet{choi2021ilvr} and \citet{dhariwal2021diffusion} without task-specific fine-tuning. Code to reproduce all experiments will be released upon acceptance.

\subsection*{Ethics statement}

This work proposes a general-purpose optimization method for image inverse problems using pretrained diffusion models. All experiments are conducted on publicly available face (FFHQ) and object (ImageNet) datasets. The method does not involve human subjects, collection of new data, or deployment in sensitive applications. As with other image reconstruction techniques, the underlying diffusion models were trained on large-scale image datasets whose curation may carry biases; we do not address dataset bias in this work.

\subsection*{AI use statement}

In this work, we have \emph{not} used generative AI tools for any of the following required-disclosure tasks: generating synthetic data, developing theoretical models or conceptual frameworks, formulating mathematical claims, assisting in the writing of proofs, proposing or refining hypotheses, designing research methodology or experiments, implementing methods, interpreting results, assisting with translation, cleaning or reformatting datasets, or supporting qualitative and thematic data analysis. All research ideas, mathematical derivations, algorithm design, and experimental conclusions are solely the work of the authors.

Additionally, we used generative AI tools (Claude via Cursor IDE) to edit and polish draft text for improved readability and English fluency. All AI-edited text was reviewed and revised by the authors to ensure accuracy. We take responsibility for the final content of this work, including all text, claims, and artifacts.

\bibliographystyle{iclr2027_conference}
\bibliography{references}

\begin{thebibliography}{34}
\providecommand{\natexlab}[1]{#1}
\providecommand{\url}[1]{\texttt{#1}}
\expandafter\ifx\csname urlstyle\endcsname\relax
  \providecommand{\doi}[1]{doi: #1}\else
  \providecommand{\doi}{doi: \begingroup \urlstyle{rm}\Url}\fi

\bibitem[Absil et~al.(2008)Absil, Mahony, and Sepulchre]{absil2008optimization}
P.-A. Absil, Robert Mahony, and Rodolphe Sepulchre.
\newblock \emph{Optimization Algorithms on Matrix Manifolds}.
\newblock Princeton University Press, 2008.
\newblock ISBN 978-0-691-13298-3.
\newblock URL \url{https://sites.uclouvain.be/absil/amsbook/}.

\bibitem[Alkhouri et~al.(2025)Alkhouri, Liang, Huang, Dai, Qu, Ravishankar, and Wang]{alkhouri2025sitcom}
Ismail Alkhouri, Shijun Liang, Cheng-Han Huang, Jimmy Dai, Qing Qu, Saiprasad Ravishankar, and Rongrong Wang.
\newblock {SITCOM}: Step-wise triple-consistent diffusion sampling for inverse problems.
\newblock In \emph{Proceedings of the 42nd International Conference on Machine Learning}, volume 267 of \emph{Proceedings of Machine Learning Research}, pp.\  1128--1158. PMLR, 2025.
\newblock URL \url{https://proceedings.mlr.press/v267/alkhouri25b.html}.

\bibitem[Anderson(1982)]{anderson1982reverse}
Brian D.~O. Anderson.
\newblock Reverse-time diffusion equation models.
\newblock \emph{Stochastic Processes and their Applications}, 12\penalty0 (3):\penalty0 313--326, 1982.
\newblock \doi{10.1016/0304-4149(82)90051-5}.
\newblock URL \url{https://doi.org/10.1016/0304-4149(82)90051-5}.

\bibitem[Boys et~al.(2024)Boys, Girolami, Pidstrigach, Reich, Mosca, and Akyildiz]{boys2024tweedie}
Benjamin Boys, Mark Girolami, Jakiw Pidstrigach, Sebastian Reich, Alan Mosca, and Omer~Deniz Akyildiz.
\newblock {Tweedie} moment projected diffusions for inverse problems.
\newblock \emph{Transactions on Machine Learning Research}, 2024.
\newblock URL \url{https://openreview.net/forum?id=4unJi0qrTE}.

\bibitem[Choi et~al.(2021)Choi, Kim, Jeong, Gwon, and Yoon]{choi2021ilvr}
Jooyoung Choi, Sungwon Kim, Yonghyun Jeong, Youngjune Gwon, and Sungroh Yoon.
\newblock {ILVR}: Conditioning method for denoising diffusion probabilistic models.
\newblock In \emph{2021 IEEE/CVF International Conference on Computer Vision (ICCV)}, pp.\  14347--14356. IEEE, 2021.
\newblock \doi{10.1109/ICCV48922.2021.01410}.
\newblock URL \url{https://doi.org/10.1109/ICCV48922.2021.01410}.

\bibitem[Chung et~al.(2022)Chung, Sim, Ryu, and Ye]{chung2022mcg}
Hyungjin Chung, Byeongsu Sim, Dohoon Ryu, and Jong~Chul Ye.
\newblock Improving diffusion models for inverse problems using manifold constraints.
\newblock In \emph{Advances in Neural Information Processing Systems}, volume~35, pp.\  25683--25696. Curran Associates, Inc., 2022.
\newblock \doi{10.52202/068431-1862}.
\newblock URL \url{https://proceedings.neurips.cc/paper_files/paper/2022/hash/a48e5877c7bf86a513950ab23b360498-Abstract-Conference.html}.

\bibitem[Chung et~al.(2023)Chung, Kim, McCann, Klasky, and Ye]{chung2023dps}
Hyungjin Chung, Jeongsol Kim, Michael~Thompson McCann, Marc~Louis Klasky, and Jong~Chul Ye.
\newblock Diffusion posterior sampling for general noisy inverse problems.
\newblock In \emph{International Conference on Learning Representations}, 2023.
\newblock URL \url{https://openreview.net/forum?id=OnD9zGAGT0k}.

\bibitem[Deng et~al.(2009)Deng, Dong, Socher, Li, Li, and Fei-Fei]{deng2009imagenet}
Jia Deng, Wei Dong, Richard Socher, Li-Jia Li, Kai Li, and Li~Fei-Fei.
\newblock {ImageNet}: A large-scale hierarchical image database.
\newblock In \emph{2009 IEEE Conference on Computer Vision and Pattern Recognition}, pp.\  248--255. IEEE Computer Society, 2009.
\newblock \doi{10.1109/CVPR.2009.5206848}.
\newblock URL \url{https://doi.org/10.1109/CVPR.2009.5206848}.

\bibitem[Dhariwal \& Nichol(2021)Dhariwal and Nichol]{dhariwal2021diffusion}
Prafulla Dhariwal and Alexander Nichol.
\newblock Diffusion models beat {GANs} on image synthesis.
\newblock In \emph{Advances in Neural Information Processing Systems}, volume~34, pp.\  8780--8794. Curran Associates, Inc., 2021.
\newblock URL \url{https://proceedings.neurips.cc/paper_files/paper/2021/hash/49ad23d1ec9fa4bd8d77d02681df5cfa-Abstract.html}.

\bibitem[Dieleman(2024)]{dieleman2024spectral}
Sander Dieleman.
\newblock Diffusion is spectral autoregression.
\newblock Sander Dieleman's blog, September 2024.
\newblock URL \url{https://sander.ai/2024/09/02/spectral-autoregression.html}.

\bibitem[Fienup(1982)]{fienup1982phase}
James~R. Fienup.
\newblock Phase retrieval algorithms: A comparison.
\newblock \emph{Applied Optics}, 21\penalty0 (15):\penalty0 2758--2769, 1982.
\newblock \doi{10.1364/AO.21.002758}.
\newblock URL \url{https://doi.org/10.1364/AO.21.002758}.

\bibitem[Ho et~al.(2020)Ho, Jain, and Abbeel]{ho2020denoising}
Jonathan Ho, Ajay Jain, and Pieter Abbeel.
\newblock Denoising diffusion probabilistic models.
\newblock In \emph{Advances in Neural Information Processing Systems}, volume~33, pp.\  6840--6851. Curran Associates, Inc., 2020.
\newblock URL \url{https://proceedings.neurips.cc/paper_files/paper/2020/hash/4c5bcfec8584af0d967f1ab10179ca4b-Abstract.html}.

\bibitem[Jia et~al.(2026)Jia, Yuan, Liu, Shen, and Wang]{jia2026weak}
Jing Jia, Wei Yuan, Sifan Liu, Liyue Shen, and Guanyang Wang.
\newblock Weak diffusion priors can still achieve strong inverse-problem performance.
\newblock In \emph{International Conference on Machine Learning}, 2026.
\newblock URL \url{https://openreview.net/forum?id=fdkSA4F0lN}.

\bibitem[Karras et~al.(2021)Karras, Laine, and Aila]{karras2019style}
Tero Karras, Samuli Laine, and Timo Aila.
\newblock A style-based generator architecture for generative adversarial networks.
\newblock \emph{IEEE Transactions on Pattern Analysis and Machine Intelligence}, 43\penalty0 (12):\penalty0 4217--4228, 2021.
\newblock \doi{10.1109/TPAMI.2020.2970919}.
\newblock URL \url{https://doi.org/10.1109/TPAMI.2020.2970919}.

\bibitem[Kawar et~al.(2022)Kawar, Elad, Ermon, and Song]{kawar2022ddrm}
Bahjat Kawar, Michael Elad, Stefano Ermon, and Jiaming Song.
\newblock Denoising diffusion restoration models.
\newblock In \emph{Advances in Neural Information Processing Systems}, volume~35, pp.\  23593--23606. Curran Associates, Inc., 2022.
\newblock \doi{10.52202/068431-1714}.
\newblock URL \url{https://proceedings.neurips.cc/paper_files/paper/2022/hash/95504595b6169131b6ed6cd72eb05616-Abstract-Conference.html}.

\bibitem[Kim et~al.(2026)Kim, Shin, and Lim]{fastdips2026}
Minwoo Kim, Seunghyeok Shin, and Hongki Lim.
\newblock {FAST-DIPS}: Adjoint-free analytic steps and hard-constrained likelihood correction for diffusion-prior inverse problems.
\newblock In \emph{International Conference on Learning Representations}, volume 2026, pp.\  104696--104730, 2026.
\newblock URL \url{https://proceedings.iclr.cc/paper_files/paper/2026/hash/ab378fb084f313a204432f0a1e697bae-Abstract-Conference.html}.

\bibitem[Laurent \& Massart(2000)Laurent and Massart]{laurent2000adaptive}
Beatrice Laurent and Pascal Massart.
\newblock Adaptive estimation of a quadratic functional by model selection.
\newblock \emph{The Annals of Statistics}, 28\penalty0 (5):\penalty0 1302--1338, 2000.
\newblock \doi{10.1214/aos/1015957395}.
\newblock URL \url{https://doi.org/10.1214/aos/1015957395}.

\bibitem[Li et~al.(2026)Li, Kwon, Liang, Alkhouri, Ravishankar, and Qu]{li2024dcdp}
Xiang Li, Soo~Min Kwon, Shijun Liang, Ismail~R. Alkhouri, Saiprasad Ravishankar, and Qing Qu.
\newblock Decoupled data consistency with diffusion purification for image restoration.
\newblock In \emph{International Conference on Computational Photography}, 2026.
\newblock URL \url{https://arxiv.org/abs/2403.06054}.

\bibitem[Liao et~al.(2026)Liao, Peng, Yueh, Sun, Chou, and Chen]{liao2026diffrgd}
Jia-Wei Liao, Li-Xuan Peng, Mei-Heng Yueh, Min Sun, Cheng-Fu Chou, and Jun-Cheng Chen.
\newblock {DiffRGD}: An inference-time diffusion guidance through {Riemannian} gradient descent.
\newblock In \emph{European Conference on Computer Vision}, 2026.
\newblock URL \url{https://media.eventhosts.cc/Conferences/ECCV2026/pdfs/9638.pdf}.

\bibitem[Lugmayr et~al.(2022)Lugmayr, Danelljan, Romero, Yu, Timofte, and Van~Gool]{lugmayr2022repaint}
Andreas Lugmayr, Martin Danelljan, Andres Romero, Fisher Yu, Radu Timofte, and Luc Van~Gool.
\newblock {RePaint}: Inpainting using denoising diffusion probabilistic models.
\newblock In \emph{2022 IEEE/CVF Conference on Computer Vision and Pattern Recognition (CVPR)}, pp.\  11451--11461. IEEE, 2022.
\newblock \doi{10.1109/CVPR52688.2022.01117}.
\newblock URL \url{https://doi.org/10.1109/CVPR52688.2022.01117}.

\bibitem[Meng \& Kabashima(2025)Meng and Kabashima]{meng2022dmps}
Xiangming Meng and Yoshiyuki Kabashima.
\newblock Diffusion model based posterior sampling for noisy linear inverse problems.
\newblock In \emph{Proceedings of the 16th Asian Conference on Machine Learning}, volume 260 of \emph{Proceedings of Machine Learning Research}, pp.\  623--638. PMLR, 2025.
\newblock URL \url{https://proceedings.mlr.press/v260/meng25a.html}.

\bibitem[Rissanen et~al.(2023)Rissanen, Heinonen, and Solin]{rissanen2022generative}
Severi Rissanen, Markus Heinonen, and Arno Solin.
\newblock Generative modelling with inverse heat dissipation.
\newblock In \emph{International Conference on Learning Representations}, 2023.
\newblock URL \url{https://openreview.net/forum?id=4PJUBT9f2Ol}.

\bibitem[Song et~al.(2021{\natexlab{a}})Song, Meng, and Ermon]{song2021denoising}
Jiaming Song, Chenlin Meng, and Stefano Ermon.
\newblock Denoising diffusion implicit models.
\newblock In \emph{International Conference on Learning Representations}, 2021{\natexlab{a}}.
\newblock URL \url{https://openreview.net/forum?id=St1giarCHLP}.

\bibitem[Song et~al.(2023)Song, Vahdat, Mardani, and Kautz]{song2023pigdm}
Jiaming Song, Arash Vahdat, Morteza Mardani, and Jan Kautz.
\newblock Pseudoinverse-guided diffusion models for inverse problems.
\newblock In \emph{International Conference on Learning Representations}, 2023.
\newblock URL \url{https://openreview.net/forum?id=9_gsMA8MRKQ}.

\bibitem[Song et~al.(2021{\natexlab{b}})Song, Sohl-Dickstein, Kingma, Kumar, Ermon, and Poole]{song2021scorebased}
Yang Song, Jascha Sohl-Dickstein, Diederik~P. Kingma, Abhishek Kumar, Stefano Ermon, and Ben Poole.
\newblock Score-based generative modeling through stochastic differential equations.
\newblock In \emph{International Conference on Learning Representations}, 2021{\natexlab{b}}.
\newblock URL \url{https://openreview.net/forum?id=PxTIG12RRHS}.

\bibitem[Tran et~al.(2021)Tran, Tran, Phung, and Hoai]{tran2021explore}
Phong Tran, Anh~Tuan Tran, Quynh Phung, and Minh Hoai.
\newblock Explore image deblurring via encoded blur kernel space.
\newblock In \emph{2021 IEEE/CVF Conference on Computer Vision and Pattern Recognition (CVPR)}, pp.\  11951--11960. IEEE, 2021.
\newblock \doi{10.1109/CVPR46437.2021.01178}.
\newblock URL \url{https://doi.org/10.1109/CVPR46437.2021.01178}.

\bibitem[Wang et~al.(2024)Wang, Zhang, Li, Wan, Chen, and Sun]{wang2024dmplug}
Hengkang Wang, Xu~Zhang, Taihui Li, Yuxiang Wan, Tiancong Chen, and Ju~Sun.
\newblock {DMPlug}: A plug-in method for solving inverse problems with diffusion models.
\newblock In \emph{Advances in Neural Information Processing Systems}, volume~37, pp.\  117881--117916. Curran Associates, Inc., 2024.
\newblock \doi{10.52202/079017-3744}.
\newblock URL \url{https://proceedings.neurips.cc/paper_files/paper/2024/hash/d5c04aa72b92c53bda5b525b60958295-Abstract-Conference.html}.

\bibitem[Wang et~al.(2023)Wang, Yu, and Zhang]{wang2023ddnm}
Yinhuai Wang, Jiwen Yu, and Jian Zhang.
\newblock Zero-shot image restoration using denoising diffusion null-space model.
\newblock In \emph{International Conference on Learning Representations}, 2023.
\newblock URL \url{https://openreview.net/forum?id=mRieQgMtNTQ}.

\bibitem[Wang et~al.(2004)Wang, Bovik, Sheikh, and Simoncelli]{wang2004image}
Zhou Wang, Alan~C. Bovik, Hamid~R. Sheikh, and Eero~P. Simoncelli.
\newblock Image quality assessment: From error visibility to structural similarity.
\newblock \emph{IEEE Transactions on Image Processing}, 13\penalty0 (4):\penalty0 600--612, 2004.
\newblock \doi{10.1109/TIP.2003.819861}.
\newblock URL \url{https://doi.org/10.1109/TIP.2003.819861}.

\bibitem[Xia et~al.(2026)Xia, Tanomkiattikun, Zhen, and Gu]{xia2026nhmc}
Yingzhi Xia, Setthakorn Tanomkiattikun, Liangli Zhen, and Zaiwang Gu.
\newblock Noise-adaptive diffusion sampling for inverse problems without task-specific tuning.
\newblock In \emph{International Conference on Learning Representations}, 2026.
\newblock URL \url{https://proceedings.iclr.cc/paper_files/paper/2026/hash/1e45dc45f1649f880130169b2271129e-Abstract-Conference.html}.

\bibitem[Yang et~al.(2024)Yang, Ding, Cai, Yu, Wang, and Shi]{yang2024dsg}
Lingxiao Yang, Shutong Ding, Yifan Cai, Jingyi Yu, Jingya Wang, and Ye~Shi.
\newblock Guidance with spherical {Gaussian} constraint for conditional diffusion.
\newblock In \emph{Proceedings of the 41st International Conference on Machine Learning}, volume 235 of \emph{Proceedings of Machine Learning Research}, pp.\  56071--56095. PMLR, 2024.
\newblock URL \url{https://proceedings.mlr.press/v235/yang24h.html}.

\bibitem[Zhang et~al.(2025)Zhang, Chu, Berner, Meng, Anandkumar, and Song]{zhang2024daps}
Bingliang Zhang, Wenda Chu, Julius Berner, Chenlin Meng, Anima Anandkumar, and Yang Song.
\newblock Improving diffusion inverse problem solving with decoupled noise annealing.
\newblock In \emph{Proceedings of the IEEE/CVF Conference on Computer Vision and Pattern Recognition}, pp.\  20895--20905, 2025.
\newblock \doi{10.1109/CVPR52734.2025.01946}.
\newblock URL \url{https://openaccess.thecvf.com/content/CVPR2025/html/Zhang_Improving_Diffusion_Inverse_Problem_Solving_with_Decoupled_Noise_Annealing_CVPR_2025_paper.html}.

\bibitem[Zhang et~al.(2026)Zhang, Liu, Yan, Xiao, and Gu]{zhan2026maprps}
Jiawei Zhang, Ziyuan Liu, Leon Yan, Zhenyu Xiao, and Yuantao Gu.
\newblock Stage-wise distortion--perception traversal in zero-shot inverse problems with diffusion models.
\newblock In \emph{International Conference on Machine Learning}, 2026.
\newblock URL \url{https://arxiv.org/abs/2605.28711}.

\bibitem[Zhang et~al.(2018)Zhang, Isola, Efros, Shechtman, and Wang]{zhang2018unreasonable}
Richard Zhang, Phillip Isola, Alexei~A. Efros, Eli Shechtman, and Oliver Wang.
\newblock The unreasonable effectiveness of deep features as a perceptual metric.
\newblock In \emph{Proceedings of the IEEE Conference on Computer Vision and Pattern Recognition}, pp.\  586--595, 2018.
\newblock URL \url{https://openaccess.thecvf.com/content_cvpr_2018/html/Zhang_The_Unreasonable_Effectiveness_CVPR_2018_paper.html}.

\end{thebibliography}

\clearpage
\appendix
\numberwithin{equation}{section}
\numberwithin{figure}{section}
\section{Theoretical Derivations and Proofs}
\label{app:derivations}

\subsection{Derivation of the Radial Potential}
\label{app:radial_potential_derivation}

A standard Gaussian vector $\ve\sim\mathcal N(\bm{0},\mI_d)$ has density $p(\ve)\propto\exp(-\|\ve\|_2^2/2)$. We decompose $\ve$ in polar coordinates as $\ve=r\cdot \vu$, where the normalized radial magnitude is $r:=\|\ve\|_2/\sqrt{d-1}>0$ and the direction is $\vu\in\mathbb S_{\sqrt{d-1}}^{d-1}:=\{\vu\in\mathbb R^d:\|\vu\|_2=\sqrt{d-1}\}$. Since $\|\ve\|_2=r\sqrt{d-1}$, the Jacobian of the polar change of variables (the spherical surface element in $d$ dimensions multiplied by the radial factor $\|\ve\|_2^{d-1}$) yields the joint density
\begin{equation}
    p(r,\vu)\propto (r\sqrt{d-1})^{d-1}\exp\!\left(-\frac{(d-1)\,r^2}{2}\right),
    \label{eq:app_polar_density}
\end{equation}
where the direction $\vu$ is uniformly distributed on $\mathbb S_{\sqrt{d-1}}^{d-1}$. Integrating over $\vu$ gives the radial marginal density
\begin{equation}
    p(r)\propto r^{d-1}\exp\!\left(-\frac{(d-1)\,r^2}{2}\right),\quad r>0.
    \label{eq:app_radial_marginal}
\end{equation}
Taking the negative log and subtracting a constant so that the minimum is zero gives the radial potential
\begin{equation}
    \Phi_d(r):=-\log p(r)+\mathrm{const}
    =\frac{d-1}{2}\bigl(r^2-1-2\log r\bigr),
    \label{eq:app_radial_potential_derivation}
\end{equation}
which attains its minimum value of zero at $r=1$ (i.e., $\|\ve\|_2=\sqrt{d-1}$). Its curvature at $r=1$ is $\Phi_d''(1)=2(d-1)$, reflecting the strength of high-dimensional concentration: larger $d$ imposes a greater cost for deviations from the typical value.

\subsection{Proof of Proposition~\ref{prop:radial_relaxation}}
\label{app:radial_relaxation_proof}

Following the notation in the main text, fix the annealing layer $t$, anchor $\bar\vx$, and direction $\vu\in\mathbb S_{\sqrt{d-1}}^{d-1}$, and write $F(r):=F_t(r;\vu)$, $J(r):=J_\lambda(r;\vu)=F(r)+\lambda_{\mathrm{rad}}\Phi_d(r)$, and $r_\lambda:=r_\lambda(\vu)$.

\begin{proof}[Proof of Proposition~\ref{prop:radial_relaxation}]\par
\noindent\textbf{(i): Existence and bounded deviation.}
Since $\log r\le r-1$, we have $r^2-1-2\log r=(r-1)^2+2(r-1-\log r)\ge(r-1)^2$, and therefore
\begin{equation}
    \Phi_d(r)\ge\frac{d-1}{2}(r-1)^2.
    \label{eq:app_phi_quadratic_lb}
\end{equation}
Moreover, $\Phi_d(r)\to+\infty$ as $r\to0^+$ or $r\to+\infty$. Together with $F(r)\ge0$, this implies $J(r)\to+\infty$. Every sublevel set at a finite level is closed and bounded, so continuity ensures that $J$ admits a global minimizer $r_\lambda\in(0,\infty)$. By the global optimality of $r_\lambda$ and $\Phi_d(1)=0$,
\[
    \lambda_{\mathrm{rad}}\Phi_d(r_\lambda)
    \le J(r_\lambda)
    \le J(1)=F(1).
\]
Combining this with Equation~\eqref{eq:app_phi_quadratic_lb} gives $\frac{\lambda_{\mathrm{rad}}(d-1)}{2}(r_\lambda-1)^2\le F(1)$, or equivalently $|r_\lambda-1|\le\sqrt{2F(1)/[\lambda_{\mathrm{rad}}(d-1)]}$. Define $\vz_t(r,\vu):=a_t\bar\vx+\sigma_t r\vu$; then $\|\vz_t(r,\vu)-\vz_t(1,\vu)\|_2=\sigma_t|r-1|\sqrt{d-1}$. Substituting the radius bound above yields the following bound on the denoiser input shift:
\begin{equation}
    \|\vz_t(r_\lambda,\vu)-\vz_t(1,\vu)\|_2\le\sigma_t\sqrt{\frac{2F(1)}{\lambda_{\mathrm{rad}}}}\,.
    \label{eq:app_input_shift_bound}
\end{equation}

\noindent\textbf{(ii): Strict relaxation gain.}
Since $\Phi_d'(1)=0$, we have $J'(1)=F'(1)=:s_t(\vu)$. If $s_t(\vu)\neq0$, then $r{=}1$ is not a stationary point of $J$. Set $\delta=-\varepsilon\,\mathrm{sign}(s_t(\vu))$ for sufficiently small $\varepsilon>0$. Then
\[
    J(1+\delta)=J(1)+s_t(\vu)\delta+o(|\delta|)
    =J(1)-\varepsilon|s_t(\vu)|+o(\varepsilon)<J(1).
\]
By the global optimality of $r_\lambda$, $J(r_\lambda)\le J(1+\delta)<J(1)=F(1)$.
\end{proof}

\subsection{Strong-Regularization Asymptotic Expansion}

\begin{corollary}[Strong-regularization asymptotics]
\label{cor:asymptotic}
Under the assumptions of Proposition~\ref{prop:radial_relaxation}, further assume that $F$ is twice continuously differentiable in a neighborhood of $r{=}1$, and let $c_\lambda:=\lambda_{\mathrm{rad}}(d-1)$ and $s_t(\vu):=F'(1)$. Fix $d$, $t$, $\bar\vx$, and $\vu$, and let $\lambda_{\mathrm{rad}}\to\infty$ (equivalently, $c_\lambda\to\infty$). Then
\begin{equation}
    r_\lambda(\vu)=1-\frac{s_t(\vu)}{2c_\lambda}+O(c_\lambda^{-2}),
    \label{eq:app_asymptotic_r}
\end{equation}
\begin{equation}
    J_\lambda(r_\lambda(\vu);\vu)=F_t(1;\vu)-\frac{s_t(\vu)^2}{4c_\lambda}+O(c_\lambda^{-2}).
    \label{eq:app_asymptotic_J}
\end{equation}
\end{corollary}

\begin{proof}[Proof]
By part (i), $r_\lambda\to1$ as $c_\lambda\to\infty$. Let $\delta_\lambda:=r_\lambda-1$. The first-order condition at an interior minimizer is
\[
    0=F'(r_\lambda)+c_\lambda\!\left[(1+\delta_\lambda)-\frac{1}{1+\delta_\lambda}\right].
\]
Part (i) gives $\delta_\lambda=O(c_\lambda^{-1/2})$, but a tighter order is needed below. Since $r_\lambda\to1$ and $F'$ is continuous near $1$, there exists a $c_\lambda$-independent constant $M>0$ such that $|F'(r_\lambda)|\le M$. Using $r_\lambda-1/r_\lambda=\delta_\lambda(r_\lambda+1)/r_\lambda$, the first-order condition gives $|\delta_\lambda|=|F'(r_\lambda)|\cdot r_\lambda/[c_\lambda(r_\lambda+1)]\le M/c_\lambda$, hence $\delta_\lambda=O(c_\lambda^{-1})$.

Expanding the bracketed term as $2\delta_\lambda+O(\delta_\lambda^2)$ and using $F'(1+\delta_\lambda)=s_t(\vu)+O(\delta_\lambda)$ gives
\[
    0=s_t(\vu)+2c_\lambda\delta_\lambda+O(c_\lambda^{-1}),
\]
which yields Equation~\eqref{eq:app_asymptotic_r}. A Taylor expansion of the objective gives
\[
    \begin{aligned}
        F(1+\delta_\lambda)
        &=F(1)+s_t(\vu)\delta_\lambda+O(\delta_\lambda^2),\\
        \lambda_{\mathrm{rad}}\Phi_d(1+\delta_\lambda)
        &=c_\lambda\delta_\lambda^2+O(c_\lambda\delta_\lambda^3).
    \end{aligned}
\]
Substituting $\delta_\lambda=-s_t(\vu)/(2c_\lambda)+O(c_\lambda^{-2})$, the measurement loss contributes $-s_t(\vu)^2/(2c_\lambda)$ and the potential contributes $s_t(\vu)^2/(4c_\lambda)$. Combining the two yields Equation~\eqref{eq:app_asymptotic_J}.
\end{proof}

Equations~\eqref{eq:app_asymptotic_r}--\eqref{eq:app_asymptotic_J} show that, under strong regularization, the optimal radius has an $O(c_\lambda^{-1})$ deviation from $1$; when $s_t(\vu)\neq0$, this corresponds to an objective decrease of order $s_t(\vu)^2/(4c_\lambda)$. Radial regularization thus limits the magnitude of radial deviations without entirely eliminating measurement-gradient-driven radial adjustment.

\clearpage
\section{Implementation Details and Experimental Setup}
\label{app:experiment_config}

\subsection{Algorithm Pseudocode}
\label{app:algorithm}

\par\medskip\noindent
\begin{minipage}{\linewidth}
\hrule height 0.8pt\vspace{4pt}
\algorithmcaption{\textbf{FANO}: Anchored Forward-Noise Optimization}
\label{alg:fano}
\vspace{2pt}\hrule height 0.4pt\vspace{4pt}
\begingroup
\renewcommand{\arraystretch}{1.12}
\noindent\begin{tabular}{@{}p{0.15\linewidth}@{}p{0.85\linewidth}@{}}
\textbf{Input} & Measurement $\vy$, forward operator $A$, denoiser $D_t$.\\
& Schedule $\{t_k,a_{t_k},\sigma_{t_k}\}_{k=1}^N$; at most $K$ inner steps per layer.\\
& Learning rates $\eta_u,\eta_r$; radial penalty $\lambda_{\mathrm{rad}}$; threshold $\tau_{\mathrm{stop}}$.\\
\textbf{Output} & Restored image $\hat\vx^N$.
\end{tabular}
\endgroup
\vspace{4pt}
\begin{algorithmic}[1]
\STATE Sample $\veps_0\sim\mathcal N(\bm{0},\mI_d)$; set $\hat\vx^0\gets D_{t_1}(\sigma_{t_1}\veps_0)$
\item[] \textit{Outer loop: annealing}
\FOR{$k=1,\dots,N$}
    \STATE $\bar\vx\gets\hat\vx^{k-1}$; sample $\veps\sim\mathcal N(\bm{0},\mI_d)$
    \STATE $\vu\gets\sqrt{d-1}\,\veps/\|\veps\|_2$; $r\gets1$
    \item[] \textit{Inner loop: polar optimization}
    \FOR{$j=1,\dots,K$}
        \STATE $\vz\gets a_{t_k}\bar\vx+\sigma_{t_k}r\vu$; $\vx\gets D_{t_k}(\vz)$
        \STATE $\rho\gets\|A(\vx)-\vy\|_2$
        \IF{$\rho\le\tau_{\mathrm{stop}}$}
            \STATE \textbf{break}
        \ENDIF
        \STATE $\vg_z\gets\nabla_{\vz}\|A(D_{t_k}(\vz))-\vy\|_2^2$
        \STATE $\tilde{\vg}_u\gets P_{T_{\vu}\mathbb S}(\sigma_{t_k}r\vg_z)$
        \STATE $g_r\gets\sigma_{t_k}\langle\vg_z,\vu\rangle+\lambda_{\mathrm{rad}}\Phi_d'(r)$
        \STATE $\vu\gets\mathrm{RiemannianAdam}(\vu,\tilde{\vg}_u,\eta_u)$; $r\gets\mathrm{Adam}(r,g_r,\eta_r)$
    \ENDFOR
    \item[] \textit{Update the anchor}
    \STATE $\hat\vx^k\gets D_{t_k}(a_{t_k}\bar\vx+\sigma_{t_k}r\vu)$
\ENDFOR
\RETURN $\hat\vx^N$
\end{algorithmic}
\vspace{4pt}\hrule height 0.8pt
\end{minipage}
\par\medskip

\subsection{Polar Gradients and Optimizers}
\label{app:optimizer_details}

\paragraph{Separation of polar gradients}
\label{app:gradient_separation}

The within-layer objective in Equation~\eqref{eq:polar_map} is
\[
    J_\lambda(r;\vu)=\ell_y\!\left(D_t(a_t\bar\vx+\sigma_t r \vu)\right)+\lambda_{\mathrm{rad}}\,\Phi_d(r).
\]
Since $\Phi_d(r)$ is independent of $\vu$, the gradient with respect to $\vu$ comes entirely from the measurement loss:
\begin{equation}
    \nabla_{\vu}J_\lambda(r;\vu)
    =\sigma_t r\,\nabla_{\vz}\ell_y(D_t(\vz))\bigg|_{\vz=a_t\bar\vx+\sigma_t r\vu}.
    \label{eq:app_angular_grad}
\end{equation}
The gradient with respect to $r$ contains contributions from both the measurement loss and the radial potential:
\begin{equation}
    \frac{\partial J_\lambda(r;\vu)}{\partial r}=\sigma_t\left\langle\nabla_{\vz}\ell_y(D_t(\vz)),\vu\right\rangle\bigg|_{\vz=a_t\bar\vx+\sigma_t r \vu}+\lambda_{\mathrm{rad}}\,\Phi_d'(r),
    \label{eq:app_radial_grad}
\end{equation}
where $\Phi_d'(r)=(d-1)(r-1/r)$. Because the angular Riemannian gradient uses the tangent-space projection~\eqref{eq:tangent_projection} to further remove the radial component along $\vu$, the updates to the two variables are geometrically separated: the angular update moves $\vu$ along a tangent direction of the sphere, while the radial update adjusts the scalar $r$. The radial regularization term $\lambda_{\mathrm{rad}}\Phi_d'(r)$ appears only in the latter.

\paragraph{Riemannian Adam for the direction variable}

Fix the outer layer $k$, let $t=t_k$, and write $\mathbb S:=\mathbb S_{\sqrt{d-1}}^{d-1}$. The angular variable $\vu$ is optimized on $\mathbb S$. Denote the variables at step $j\ge1$ by $(r_j,\vu_j)$. The denoiser input and its input-space gradient are, respectively,
\[
    \vz_j:=a_t\bar\vx+\sigma_t r_j\vu_j,
    \qquad
    \vg_{z,j}:=\nabla_{\vz}\ell_y(D_t(\vz))\big|_{\vz=\vz_j}.
\]
The corresponding Euclidean angular gradient is $\vg_{u,j}:=\sigma_t r_j\vg_{z,j}=\nabla_{\vu}J_\lambda(r_j;\vu)\big|_{\vu=\vu_j}$. The update rules are as follows:

\noindent\textbf{(i)}~\emph{Tangent-space projection.} Project the Euclidean gradient onto the tangent space at $\vu_j$:
\begin{equation}
    \tilde{\vg}_{u,j}=\vg_{u,j}-\frac{\langle \vg_{u,j},\vu_j\rangle}{\|\vu_j\|_2^2}\,\vu_j.
    \label{eq:tangent_projection}
\end{equation}

\noindent\textbf{(ii)}~\emph{Adam moment updates.} Maintain first- and second-moment estimates in the tangent space:
\begin{equation}
    \begin{aligned}
    \bm{m}_j&=\beta_1 \bm{m}_{j-1}+(1-\beta_1)\tilde{\vg}_{u,j},\\
    \bm{v}_j&=\beta_2 \bm{v}_{j-1}+(1-\beta_2)\tilde{\vg}_{u,j}\odot\tilde{\vg}_{u,j}.
    \end{aligned}
\end{equation}
Apply bias correction $\hat{\bm{m}}_j=\bm{m}_j/(1-\beta_1^j)$ and $\hat{\bm{v}}_j=\bm{v}_j/(1-\beta_2^j)$. After computing the preconditioned update direction, project it onto the tangent space once more:
\begin{equation}
    \bm{\Delta}_j=P_{T_{\vu_j}\mathbb S}\!\left(\frac{\hat{\bm{m}}_j}{\sqrt{\hat{\bm{v}}_j}+\epsilon}\right).
\end{equation}

\noindent\textbf{(iii)}~\emph{Retraction.} Retract the tangent update onto the sphere:
\begin{equation}
    \vu_{j+1}=\sqrt{d{-}1}\,\frac{\vu_j-\eta_u\bm{\Delta}_j}{\|\vu_j-\eta_u\bm{\Delta}_j\|_2}.
    \label{eq:retraction_appendix}
\end{equation}

\noindent\textbf{(iv)}~\emph{Momentum transport.} After the update, reproject the first moment onto the tangent space at the new point:
\[
    \bm{m}_j\leftarrow P_{T_{\vu_{j+1}}\mathbb S}(\bm{m}_j).
\]
This keeps the momentum on the manifold. The default hyperparameters are $\beta_1=0.9$, $\beta_2=0.999$, and $\epsilon=10^{-8}$.

\paragraph{Adam for the radius variable and positivity safeguards}

The radial scalar $r>0$ is updated with standard Adam. If the updated value satisfies $r'\le r_{\min}$ (with the default $r_{\min}=10^{-6}$), a backtracking search repeatedly halves the step size until $r'>r_{\min}$ or the maximum number of backtracking steps is reached. This mechanism keeps $r$ positive throughout optimization, ensuring that the potential $\Phi_d(r)$ has a well-defined $\log r$ term. Backtracking was never triggered on any task in our experiments, indicating that $r$ does not approach zero along normal optimization trajectories.

\subsection{Statistical Motivation for Early Stopping}
\label{app:early_stopping}

For a fixed outer layer $k$, within-layer optimization uses the measurement residual for early stopping, terminating when $\|A(D_{t_k}(\vz_j))-\vy\|_2\le\tau_{\mathrm{stop}}$. The threshold is set to
\begin{equation}
    \tau_{\mathrm{stop}}:=\sigma_y\sqrt{m},
    \label{eq:app_tau_stop}
\end{equation}
where $\sigma_y$ is the standard deviation of the measurement noise and $m$ is the measurement dimension. This choice is motivated by the concentration of additive Gaussian noise: if $\vxi\sim\mathcal N(\bm{0},\sigma_y^2 \mI_m)$, then $\mathbb E[\|\vxi\|_2^2]=m\sigma_y^2$, and concentration inequalities for the $\chi^2$ distribution imply $\|\vxi\|_2/(\sigma_y\sqrt{m})\approx 1$, with deviations of order $O(1/\sqrt{m})$. Thus, $\sigma_y\sqrt{m}$ is the typical norm of the measurement noise: a residual below this level indicates that the optimizer is beginning to fit the noise itself, so further optimization harms generalization. In the noiseless case $\sigma_y=0$, we have $\tau_{\mathrm{stop}}=0$, and the criterion reduces to a fixed iteration budget.

\subsection{Task-Specific Degradation Parameters}
\label{app:task_details}

Table~\ref{tab:app_task_params} lists the degradation-operator parameters for each task.

\begin{table}[!t]
\centering
\caption{Degradation-operator parameters for each task.}
\label{tab:app_task_params}
\footnotesize
\begin{tabular}{lll}
\toprule
\textbf{Task} & \textbf{Type} & \textbf{Parameters} \\
\midrule
SR $\times 4$ & Linear & Bicubic, factor $4$ \\
Gauss.\ deblur & Linear & $61{\times}61$ Gaussian, $\sigma_{\rm blur}{=}3.0$ \\
Motion deblur & Linear & $61{\times}61$ motion, intensity $0.5$ \\
Inpaint (box) & Linear & $128{\times}128$ random mask \\
Inpaint (rand.) & Linear & Pixel mask prob.\ $0.7$ \\
Phase retrieval & Nonlinear & Oversampling ratio $1.5$ \\
HDR & Nonlinear & Clipping factor $2$ \\
NL deblur & Nonlinear & BKSE varying kernel \\
\bottomrule
\end{tabular}
\end{table}

\subsection{Task-Specific Hyperparameters for FANO}
\label{app:hyperparameters}

Table~\ref{tab:app_fano_params} lists the number of annealing layers $N$, the number of inner steps $K$, and the radial regularization coefficient $\lambda_{\mathrm{rad}}$ used for each task in the main FANO experiments. The same configurations are used for FFHQ and ImageNet. The direction is optimized with Riemannian Adam and the radius with Adam, both using learning rates $\eta_u=\eta_r=0.02$, $\beta_1=0.9$, and $\beta_2=0.999$. The annealing schedule $\{t_k\}_{k=1}^N$ is sampled at uniform intervals from a $1000$-step VP diffusion process. The forward coefficients at each layer are $a_{t_k}=\sqrt{\bar\alpha_{t_k}}$ and $\sigma_{t_k}=\sqrt{1-\bar\alpha_{t_k}}$, where $\bar\alpha_t$ denotes the cumulative signal-to-noise ratio schedule of VP diffusion. All tasks use the same scheduling procedure.

\begin{table}[!t]
\centering
\caption{Annealing configurations and radial regularization coefficients used in the main FANO experiments on FFHQ and ImageNet ($\sigma_y=0.05$).}
\label{tab:app_fano_params}
\footnotesize
\begin{tabular}{lccc}
\toprule
\textbf{Task} & \textbf{Annealing layers $N$} & \textbf{Inner steps $K$} & $\lambda_{\mathrm{rad}}$ \\
\midrule
SR $\times 4$ & 50 & 30 & $0$ \\
Gauss.\ deblur & 50 & 30 & $0.0015$ \\
Motion deblur & 50 & 30 & $0.01$ \\
Inpaint (box) & 20 & 20 & $0$ \\
Inpaint (rand.) & 20 & 30 & $1$ \\
Phase retrieval & 50 & 30 & $0.01$ \\
HDR & 50 & 30 & $0.0015$ \\
NL deblur & 50 & 30 & $0.0015$ \\
\bottomrule
\end{tabular}
\end{table}

\section{Ablation and Parameter Sensitivity Analysis}
\label{app:ablation}

\subsection{A Tuning-Free Baseline: FANO-Sphere}

When task-specific tuning of $\lambda_{\mathrm{rad}}$ is undesirable, FANO-Sphere with $r\equiv1$ provides a tuning-free default. Table~\ref{tab:app_sphere_baseline} compares its PSNR with the two most closely related annealed baselines in Table~\ref{tab:main_results_005}, SITCOM and DAPS. Among the $16$ dataset--task combinations, FANO-Sphere outperforms both SITCOM and DAPS in $15$; the sole exception is ImageNet box inpainting, where DAPS leads by $0.88$~dB. Its advantage over SITCOM is particularly pronounced on nonlinear tasks: 2.85~dB on FFHQ and 3.37~dB on ImageNet for phase retrieval, and 0.93--1.05~dB for HDR. The default FANO configuration in the main text, which jointly optimizes $r,\vu$, provides a further improvement of approximately $0.8$~dB on HDR.

\begin{table}[!t]
\centering
\caption{PSNR comparison of FANO-Sphere ($r\equiv1$, without tuning $\lambda_{\mathrm{rad}}$) against annealed baselines ($\sigma_y=0.05$). The best result for each combination is in \textbf{bold}.}
\label{tab:app_sphere_baseline}
\footnotesize
\setlength{\tabcolsep}{3.8pt}
\begin{tabular}{l ccc ccc}
\toprule
& \multicolumn{3}{c}{\textbf{FFHQ}} & \multicolumn{3}{c}{\textbf{ImageNet}} \\
\cmidrule(lr){2-4} \cmidrule(lr){5-7}
\textbf{Task} & SITCOM & DAPS & \textbf{Sphere} & SITCOM & DAPS & \textbf{Sphere} \\
\midrule
SR $\times 4$     & 30.55 & 29.35 & \textbf{30.57} & 27.09 & 25.47 & \textbf{27.14} \\
Gauss.\ deblur    & 29.89 & 29.61 & \textbf{30.66} & 26.38 & 25.88 & \textbf{27.01} \\
Motion deblur     & 32.37 & 31.65 & \textbf{33.03} & 29.38 & 28.65 & \textbf{29.89} \\
Inpaint (box)     & 24.47 & 24.32 & \textbf{25.14} & 19.62 & \textbf{21.05} & 20.17 \\
Inpaint (rand.)   & 32.40 & 30.72 & \textbf{34.06} & 28.90 & 27.35 & \textbf{29.90} \\
Phase retrieval   & 29.21 & 30.09 & \textbf{32.06} & 20.49 & 22.94 & \textbf{23.86} \\
HDR               & 28.13 & 27.48 & \textbf{29.06} & 25.75 & 26.07 & \textbf{26.80} \\
NL deblur         & 29.31 & 28.53 & \textbf{30.72} & 28.47 & 27.25 & \textbf{29.16} \\
\bottomrule
\end{tabular}
\end{table}

\subsection{Sensitivity to the Hyperparameter $\lambda_{\mathrm{rad}}$}
\label{app:lambda_sensitivity}

\begin{figure}[!t]
\centering
\includegraphics[width=\textwidth]{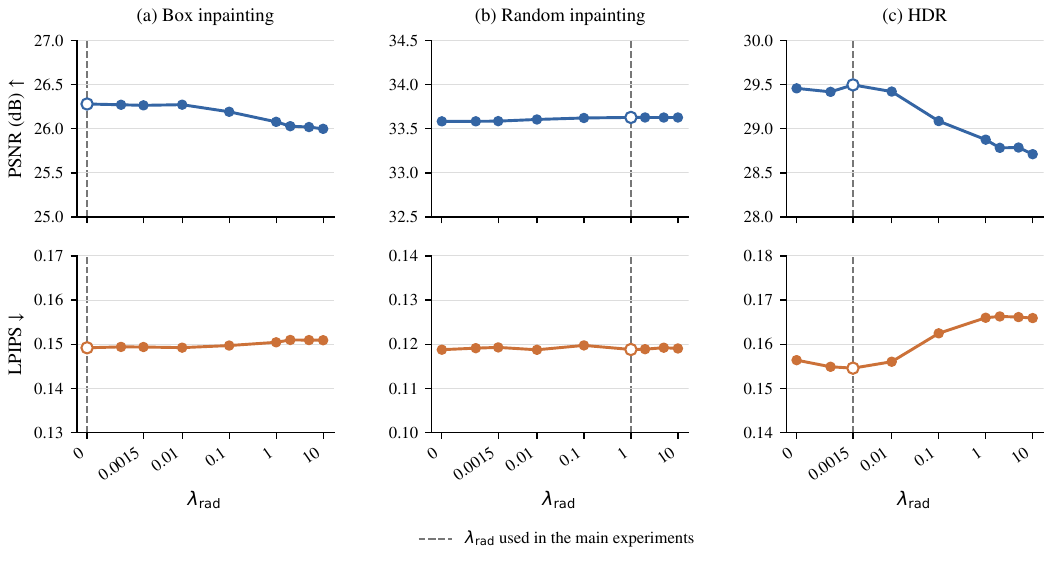}
\caption{Sensitivity to $\lambda_{\mathrm{rad}}$ on three representative tasks (the first $30$ FFHQ images, $\sigma_y{=}0.05$). The top row shows PSNR and the bottom row shows LPIPS (VGG16), both averaged over images. Dashed lines and open circles mark the default value used for each task in the main text.}
\label{fig:lambda_sensitivity}
\end{figure}

We assess the sensitivity of reconstruction quality to the radial regularization coefficient $\lambda_{\mathrm{rad}}$ through a parameter sweep. For box inpainting, random inpainting, and HDR, we use the first $30$ FFHQ images to sweep $\lambda_{\mathrm{rad}}\in\{0,0.0005,0.0015,0.01,0.1,1,2,5,10\}$. Each task uses the annealing budget in Table~\ref{tab:app_fano_params}, with the remaining hyperparameters fixed at $\eta_u{=}\eta_r{=}0.02$, $\sigma_y{=}0.05$, and $\mathrm{seed}{=}42$. Within each task, all coefficients share the same images, measurements, and initializations (Figure~\ref{fig:lambda_sensitivity}).

The curves for box inpainting and random inpainting are relatively flat. Across the full sweep, box inpainting varies by $0.28$~dB in PSNR and $0.0018$ in LPIPS, with the default $\lambda_{\mathrm{rad}}{=}0$ achieving both the highest PSNR and the lowest LPIPS. Random inpainting varies by at most $0.05$~dB in PSNR and $0.0010$ in LPIPS; the default $\lambda_{\mathrm{rad}}{=}1$ achieves the highest PSNR, and its LPIPS differs from the lowest value in the sweep by less than $0.0001$.

HDR is more sensitive to radial regularization: the default $\lambda_{\mathrm{rad}}{=}0.0015$ achieves both the highest PSNR and the lowest LPIPS on the sweep grid. Increasing the coefficient to $0.1$ or above reduces PSNR by at least $0.41$~dB and increases LPIPS by at least $0.0079$, with stronger regularization further limiting reconstruction quality. This is consistent with the role of radial freedom in nonlinear tasks (see Section~\ref{subsec:diagnosis}): when the measurement operator is strongly nonlinear, the optimizer needs moderate radial flexibility to match the measurement, and excessive radial constraints limit this ability. These results use the FFHQ30 subset and a single random seed to illustrate sensitivity trends; the configurations used for the full-dataset experiments are listed in Table~\ref{tab:app_fano_params}.

\section{Computational Cost and Efficiency Analysis}
\label{app:budget_analysis}

Do FANO's reconstruction gains arise from the method itself rather than greater computational cost? We address this question through three comparisons: (1) a direct comparison with the most closely related method, SITCOM, under matched NFE budgets to isolate the method's contribution; (2) a comparison of NFE and GPU memory efficiency with INO, which also performs spherical noise optimization; and (3) a comparison of actual runtime with DAPS.

\subsection{Matched-Budget Comparison with SITCOM}

SITCOM and FANO share the same algorithmic structure of annealing layers with within-layer optimization, making SITCOM the most directly comparable baseline in the main text. Table~\ref{tab:main_results_005} uses the $N=20$ configuration recommended in the SITCOM paper, whereas FANO defaults to $N=50, K=30$, resulting in different NFE budgets. To remove this confounding factor, Table~\ref{tab:app_sitcom_fano_budget} adjusts both the number of SITCOM annealing layers and its inner steps to match FANO (denoted SITCOM-TB), giving the two methods the same maximum NFE budget. Both methods use the residual-based early-stopping criterion in Equation~\eqref{eq:gaussian_tau_stop}.

\begin{table*}[!t]
\centering
\caption{Matched-budget comparison of SITCOM-TB and FANO at $\sigma_y=0.05$. For each task, both methods use the same number of annealing layers $N$, inner steps $K$, residual-based early-stopping criterion, and number of candidate runs. The best results are in \textbf{bold}.}
\label{tab:app_sitcom_fano_budget}
\footnotesize
\begin{tabular}{ll ccc ccc}
\toprule
& & \multicolumn{3}{c}{\textbf{FFHQ}} & \multicolumn{3}{c}{\textbf{ImageNet}} \\
\cmidrule(lr){3-5} \cmidrule(lr){6-8}
\textbf{Task} & \textbf{Method} & PSNR$\uparrow$ & SSIM$\uparrow$ & LPIPS$\downarrow$ & PSNR$\uparrow$ & SSIM$\uparrow$ & LPIPS$\downarrow$ \\
\midrule
\multirow{2}{*}{SR $\times 4$}
& SITCOM-TB & 30.61 & 0.869 & \textbf{0.226} & \textbf{27.11} & \textbf{0.759} & \textbf{0.311} \\
& FANO     & \textbf{30.68} & \textbf{0.872} & 0.232 & 27.03 & 0.755 & 0.321 \\
\midrule
\multirow{2}{*}{Gauss.\ deblur}
& SITCOM-TB & 30.09 & 0.854 & 0.244 & 26.56 & 0.732 & 0.337 \\
& FANO     & \textbf{30.65} & \textbf{0.868} & \textbf{0.225} & \textbf{27.01} & \textbf{0.746} & \textbf{0.317} \\
\midrule
\multirow{2}{*}{Motion deblur}
& SITCOM-TB & 32.62 & 0.902 & 0.187 & 29.68 & 0.838 & 0.231 \\
& FANO     & \textbf{33.04} & \textbf{0.909} & \textbf{0.167} & \textbf{29.88} & \textbf{0.840} & \textbf{0.225} \\
\midrule
\multirow{2}{*}{Inpaint (box)}
& SITCOM-TB & 24.47 & 0.737 & 0.245 & 19.62 & 0.699 & 0.256 \\
& FANO     & \textbf{25.33} & \textbf{0.866} & \textbf{0.156} & \textbf{20.40} & \textbf{0.786} & \textbf{0.206} \\
\midrule
\multirow{2}{*}{Inpaint (rand.)}
& SITCOM-TB & 32.38 & 0.863 & 0.182 & 28.89 & 0.814 & 0.181 \\
& FANO     & \textbf{34.07} & \textbf{0.931} & \textbf{0.120} & \textbf{29.94} & \textbf{0.864} & \textbf{0.157} \\
\midrule
\multirow{2}{*}{Phase retrieval}
& SITCOM-TB & 30.47 & 0.868 & 0.232 & 23.36 & \textbf{0.577} & 0.402 \\
& FANO     & \textbf{31.98} & \textbf{0.877} & \textbf{0.187} & \textbf{23.83} & 0.567 & \textbf{0.388} \\
\midrule
\multirow{2}{*}{HDR}
& SITCOM-TB & 28.03 & 0.809 & 0.216 & 25.81 & 0.765 & 0.237 \\
& FANO     & \textbf{29.88} & \textbf{0.878} & \textbf{0.157} & \textbf{27.32} & \textbf{0.832} & \textbf{0.196} \\
\midrule
\multirow{2}{*}{NL deblur}
& SITCOM-TB & 29.68 & 0.815 & 0.237 & 29.00 & \textbf{0.820} & \textbf{0.197} \\
& FANO     & \textbf{30.83} & \textbf{0.873} & \textbf{0.183} & \textbf{29.09} & 0.819 & 0.222 \\
\bottomrule
\end{tabular}
\end{table*}

Under the same annealing budget, FANO outperforms SITCOM-TB in FFHQ PSNR on all eight tasks, with particularly pronounced gains on inpainting and nonlinear tasks: $0.86$\,dB for box inpainting and $1.51$\,dB for phase retrieval on FFHQ, consistent with the analysis in Section~\ref{subsec:diagnosis}. On ImageNet, its PSNR for SR $\times4$ is lower by $0.08$\,dB and its LPIPS for nonlinear deblurring is higher by $0.025$, with trade-offs on individual metrics more apparent for linear degradation tasks. Overall, FANO's gains on nonlinear tasks arise from using $\ve$ as the search variable rather than from an additional NFE budget.

\subsection{Computational Cost Comparison with INO}

Both FANO and INO \citep{jia2026weak} optimize noise variables on a sphere, but use different inference frameworks: FANO progressively refines the estimate across annealing layers, requiring gradients through only a single denoiser call per layer, whereas INO treats the diffusion model as a black box and optimizes the initial noise of a three-step DDIM chain end to end. FANO uses a total NFE of $N \times K = 50 \times 30 = 1500$, whereas each of INO's $1000$ AdamSphere steps requires $3$ DDIM forward steps, for a total NFE of $3000$, twice that of FANO. In terms of GPU memory, INO must backpropagate through the full DDIM chain and reaches approximately 5.34\,GiB of peak allocated GPU memory, compared with only 2.30\,GiB for FANO, less than half as much. Despite its substantially lower NFE and memory usage, across the $30$ metrics reported by both methods, FANO achieves better mean performance on $29$ (Table~\ref{tab:main_results_005}). Under these experimental conditions, spherical search within a layerwise annealing framework outperforms end-to-end black-box optimization in both reconstruction quality and computational cost.

\subsection{Runtime Comparison with DAPS}

Table~\ref{tab:app_daps_runtime} reports the average core reconstruction time of FANO and DAPS \citep{zhang2024daps} on FFHQ, with all configurations executed sequentially on the same NVIDIA A100-PCIE-40GB GPU. DAPS uses its recommended 1k configuration, while FANO uses the task-specific annealing budgets, with $N=50,K=30$ for most tasks. For nonlinear deblurring, DAPS uses $200$ Langevin updates per layer, compared with $100$ for the other tasks. The nonlinear blur operator also requires calls to the BKSE network, making this task substantially more time-consuming than linear tasks.

\begin{table}[!t]
\centering
\caption{Average core reconstruction time (seconds) of FANO and DAPS on FFHQ, executed sequentially on a single A100.}
\label{tab:app_daps_runtime}
\footnotesize
\begin{tabular}{lrr}
\toprule
\textbf{Task} & \textbf{DAPS} & \textbf{FANO} \\
\midrule
SR $\times 4$     & 83  & 74  \\
Gauss.\ deblur    & 68  & 88  \\
Motion deblur     & 68  & 88  \\
Inpaint (box)     & 46  & 86  \\
Inpaint (rand.)   & 46  & 34  \\
Phase retrieval   & 107 & 87  \\
HDR               & 57  & 90  \\
NL deblur         & 606 & 101 \\
\bottomrule
\end{tabular}
\end{table}

\subsection{Total Budget and Budget Allocation}
\label{app:nfe_scaling}

\begin{figure}[!t]
\centering
\includegraphics[width=0.8\columnwidth]{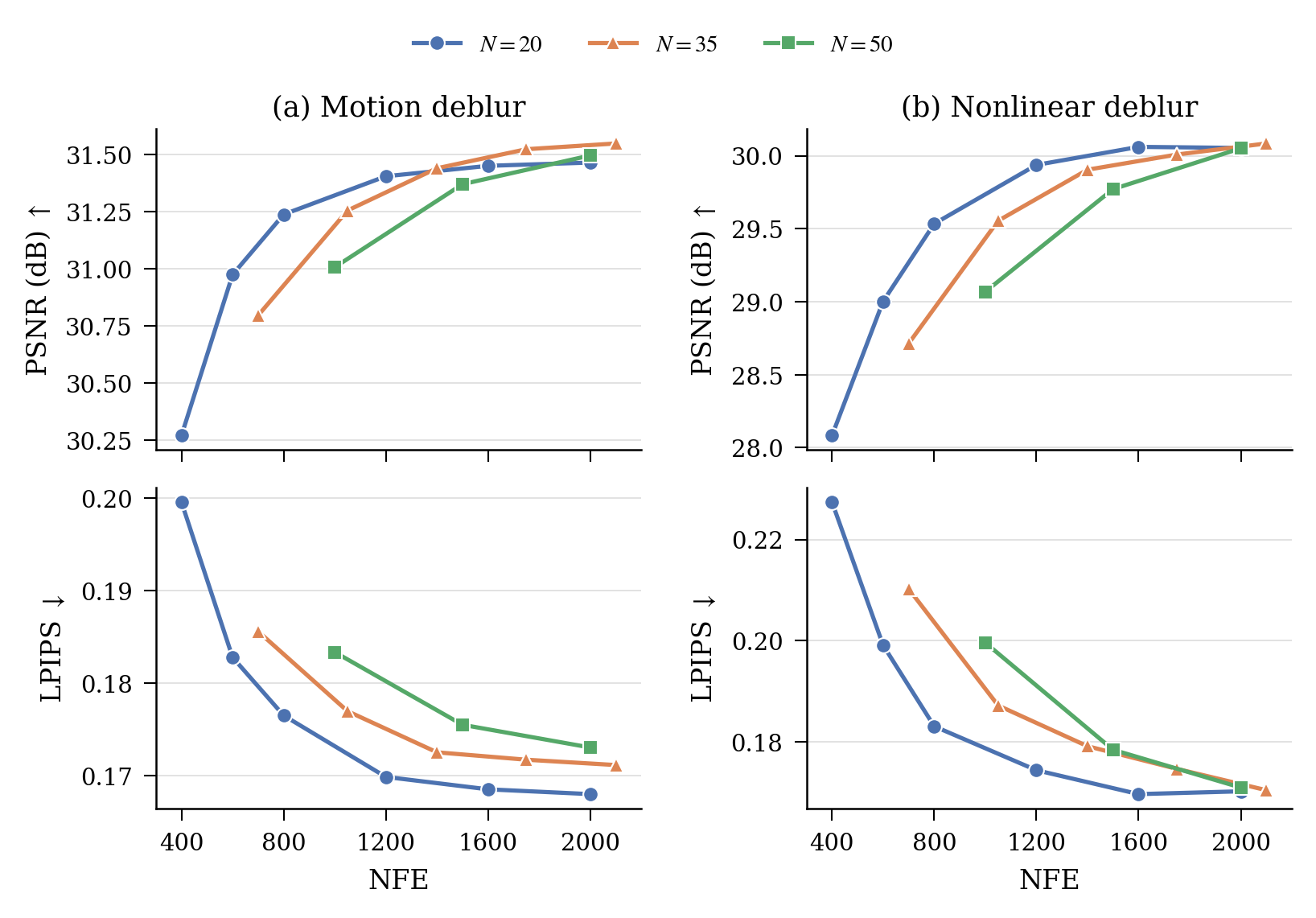}
\caption{FANO reconstruction performance under different NFE budgets. The left column shows motion deblurring and the right column nonlinear deblurring; the top row shows PSNR ($\uparrow$) and the bottom row LPIPS ($\downarrow$). Each curve corresponds to a number of annealing layers $N$, with total NFE ($=N \times K$) on the horizontal axis.}
\label{fig:nfe_scaling}
\end{figure}

This section presents two sets of budget experiments. Figure~\ref{fig:nfe_scaling} sweeps FANO's annealing budgets on the FFHQ10 subset, and Table~\ref{tab:n50_vs_n20} compares two annealing configurations on the full datasets.

Figure~\ref{fig:nfe_scaling} shows FANO's performance on one linear task (motion deblurring) and one nonlinear task (nonlinear deblurring) for different numbers of annealing layers $N$ and inner steps $K$, as a function of total NFE ($=N \times K$).

\paragraph{Monotonic improvement for fixed $N$}
For each fixed $N$, increasing $K$ monotonically improves both PSNR and LPIPS, with diminishing marginal gains as NFE increases.

\paragraph{Convergence of curves for different $N$}
The three curves corresponding to different $N$ tend to converge in the high-NFE regime (at NFE$\approx$2000, LPIPS differences are $<$0.005). This indicates that FANO's final reconstruction quality depends primarily on the total computational budget and is relatively insensitive to its allocation between $N$ and $K$. Notably, curves for larger $N$ retain greater curvature at high NFE, suggesting that further increasing $K$ may allow larger-$N$ configurations to match or exceed the asymptotic performance of smaller-$N$ configurations.

\paragraph{Trade-off between annealing layers and inner steps at a fixed budget}
Table~\ref{tab:n50_vs_n20} uses the same maximum NFE budget of $1500$ to compare $N=50,K=30$ (more layers with fewer steps per layer) and $N=20,K=75$ (fewer layers with more steps per layer). Linear tasks are insensitive to this allocation, with PSNR differences of at most $0.14$\,dB between the two configurations. The three nonlinear tasks respond differently: $N=50$ improves PSNR by $0.82$--$1.06$\,dB for phase retrieval and $0.11$--$0.27$\,dB for HDR, whereas $N=20$ improves PSNR by $0.18$--$0.21$\,dB for nonlinear deblurring and consistently achieves better LPIPS. The gains on phase retrieval substantially outweigh the losses on nonlinear deblurring, so Table~\ref{tab:main_results_005} adopts $N=50, K=30$ as the default configuration. Table~\ref{tab:n50_vs_n20} omits box inpainting and random inpainting; the budgets used for these two tasks in the main experiments are listed in Table~\ref{tab:app_fano_params}.

\begin{table*}[!t]
\centering
\caption{Results for two FANO configurations on FFHQ100 and ImageNet100 ($\sigma_y=0.05$): $N=50,K=30$ and $N=20,K=75$, both with a maximum NFE of $1500$. The best metric for each task and dataset is in \textbf{bold}.}
\label{tab:n50_vs_n20}
\footnotesize
\begin{tabular}{ll ccc ccc}
\toprule
& & \multicolumn{3}{c}{\textbf{FFHQ}} & \multicolumn{3}{c}{\textbf{ImageNet}} \\
\cmidrule(lr){3-5} \cmidrule(lr){6-8}
\textbf{Task} & \textbf{Config} & PSNR$\uparrow$ & SSIM$\uparrow$ & LPIPS$\downarrow$ & PSNR$\uparrow$ & SSIM$\uparrow$ & LPIPS$\downarrow$ \\
\midrule
\multirow{2}{*}{SR $\times 4$}
& $N\!=\!50, K\!=\!30$ & \textbf{30.68} & \textbf{0.872} & 0.232 & 27.03 & 0.755 & 0.321 \\
& $N\!=\!20, K\!=\!75$ & 30.64 & 0.869 & \textbf{0.226} & \textbf{27.10} & \textbf{0.756} & \textbf{0.315} \\
\midrule
\multirow{2}{*}{Gauss.\ deblur}
& $N\!=\!50, K\!=\!30$ & \textbf{30.65} & \textbf{0.868} & 0.225 & 27.01 & 0.746 & 0.317 \\
& $N\!=\!20, K\!=\!75$ & 30.55 & 0.866 & \textbf{0.217} & \textbf{27.08} & \textbf{0.750} & \textbf{0.306} \\
\midrule
\multirow{2}{*}{Motion deblur}
& $N\!=\!50, K\!=\!30$ & \textbf{33.04} & \textbf{0.909} & 0.167 & 29.88 & 0.840 & 0.225 \\
& $N\!=\!20, K\!=\!75$ & 33.01 & 0.908 & \textbf{0.161} & \textbf{30.02} & \textbf{0.845} & \textbf{0.208} \\
\midrule
\multirow{2}{*}{Phase retrieval}
& $N\!=\!50, K\!=\!30$ & \textbf{31.98} & \textbf{0.877} & \textbf{0.187} & \textbf{23.83} & \textbf{0.567} & \textbf{0.388} \\
& $N\!=\!20, K\!=\!75$ & 31.16 & 0.852 & 0.199 & 22.77 & 0.526 & 0.410 \\
\midrule
\multirow{2}{*}{HDR}
& $N\!=\!50, K\!=\!30$ & \textbf{29.88} & \textbf{0.878} & \textbf{0.157} & \textbf{27.32} & \textbf{0.832} & 0.196 \\
& $N\!=\!20, K\!=\!75$ & 29.61 & 0.868 & 0.162 & 27.21 & 0.827 & \textbf{0.182} \\
\midrule
\multirow{2}{*}{NL deblur}
& $N\!=\!50, K\!=\!30$ & 30.83 & 0.873 & 0.183 & 29.09 & 0.819 & 0.222 \\
& $N\!=\!20, K\!=\!75$ & \textbf{31.01} & \textbf{0.874} & \textbf{0.173} & \textbf{29.30} & \textbf{0.828} & \textbf{0.186} \\
\bottomrule
\end{tabular}
\end{table*}

\section{Extended Experiments}

\subsection{Noise Robustness}
\label{app:noise_robustness}

\begin{figure}[!t]
\centering
\includegraphics[width=0.8\columnwidth]{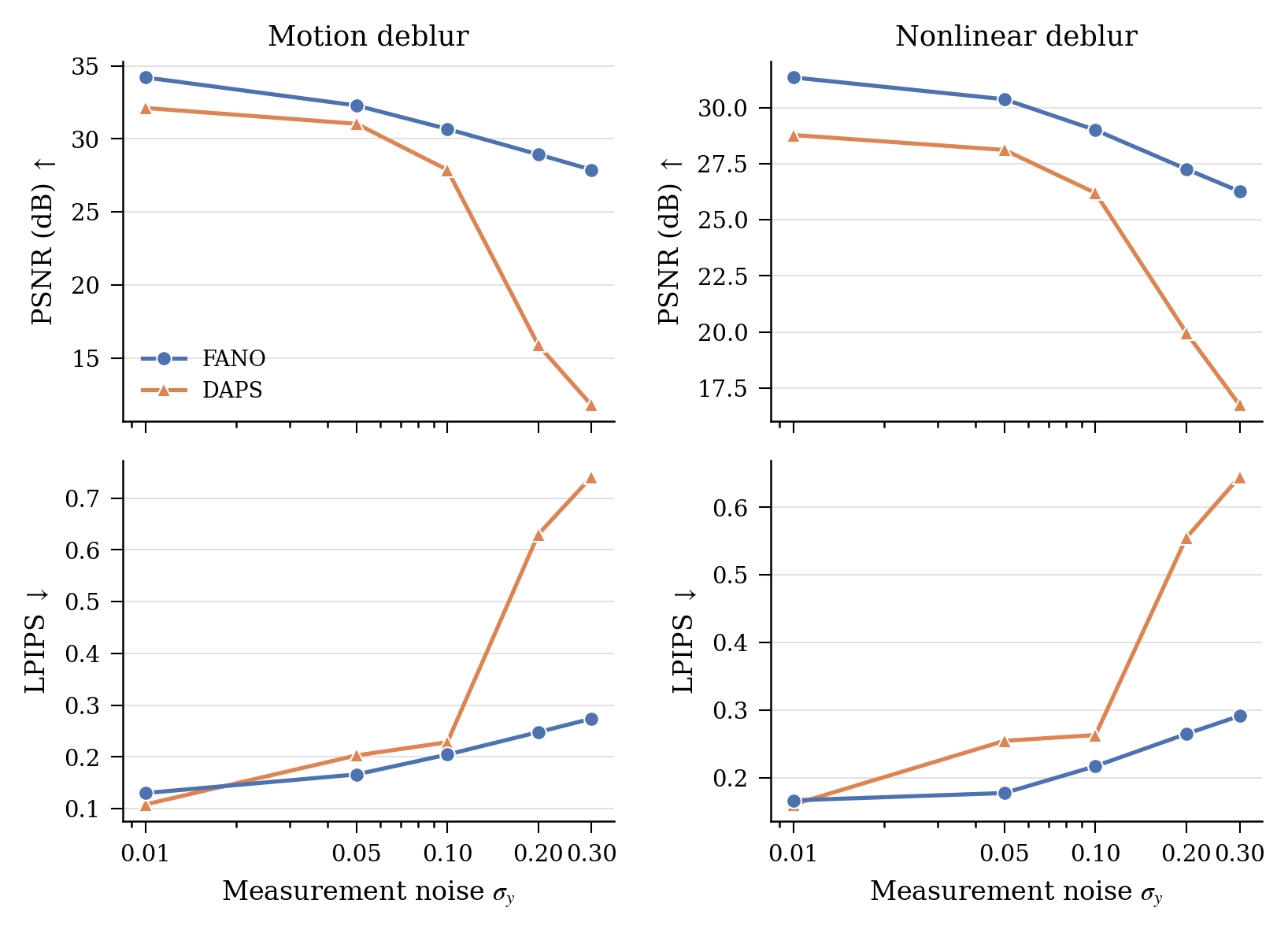}
\caption{Reconstruction performance of FANO and DAPS at different measurement noise levels $\sigma_y$. The left column shows motion deblurring and the right column nonlinear deblurring; the top row shows PSNR ($\uparrow$) and the bottom row LPIPS ($\downarrow$). FANO uses $N=50,K=30$ and DAPS uses its recommended 1k configuration. All results are averaged over 20 FFHQ images.}
\label{fig:noise_robustness}
\end{figure}

Figure~\ref{fig:noise_robustness} compares FANO and DAPS on motion deblurring and nonlinear deblurring as the measurement noise $\sigma_y$ increases from 0.01 to 0.30, using 20 FFHQ images. Both metrics for FANO degrade smoothly and monotonically: PSNR decreases from 34.21\,dB to 27.89\,dB for motion deblurring and from 31.34\,dB to 26.27\,dB for nonlinear deblurring, without abrupt changes or collapse. By contrast, DAPS performs relatively close to FANO for $\sigma_y \le 0.10$ but degrades sharply for $\sigma_y \ge 0.20$: its motion deblurring PSNR drops from 27.88\,dB at $\sigma_y=0.10$ to 15.88\,dB at $\sigma_y=0.20$, while LPIPS increases from 0.228 to 0.630.

\subsection{Phase Retrieval without Oversampling: Compensation by Generative Priors}
\label{app:pr_no_oversampling}

\begin{figure*}[!t]
\centering
\begin{minipage}[t]{0.48\textwidth}
\centering
\includegraphics[width=\linewidth]{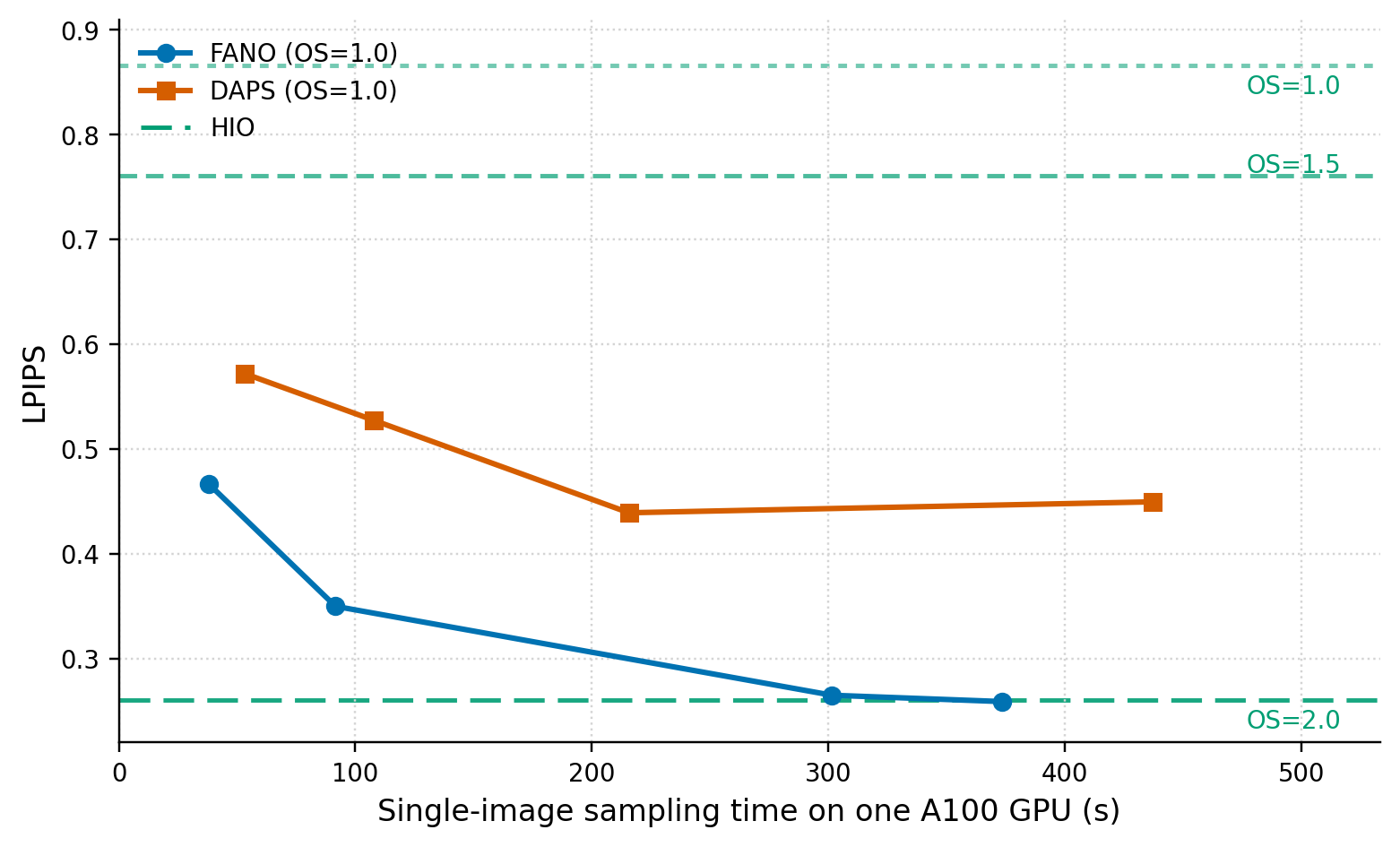}
\smallskip
\textbf{(a)} Time--LPIPS trade-off.
\end{minipage}\hfill
\begin{minipage}[t]{0.48\textwidth}
\centering
\includegraphics[width=\linewidth]{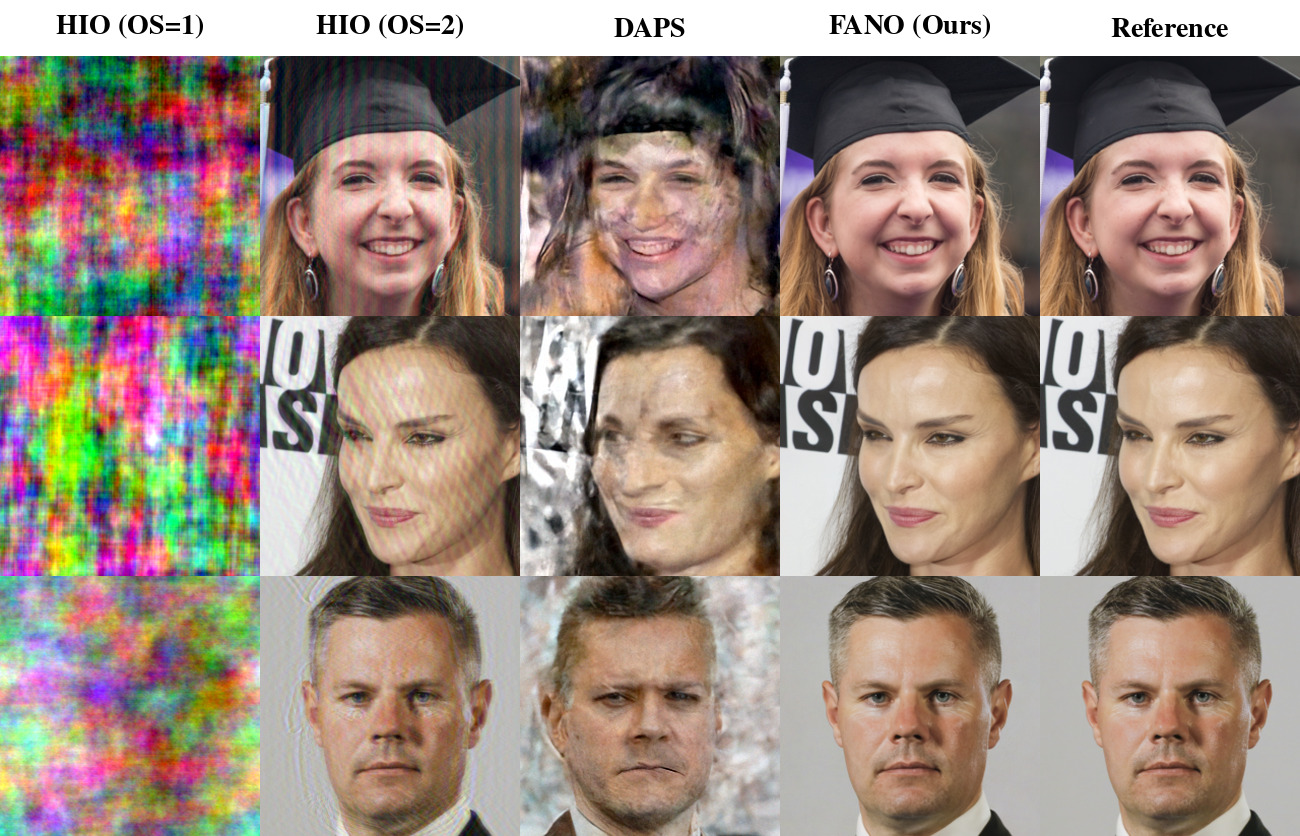}
\smallskip
\textbf{(b)} Visual comparison at $\text{OS}=1$.
\end{minipage}
\caption{Phase retrieval without oversampling ($\text{OS}=1$, noiseless) on 100 FFHQ images. (a) LPIPS as a function of per-image wall-clock time on a single A100: solid curves show FANO and DAPS under varying compute budgets; dashed lines mark HIO at three oversampling rates. (b) Three representative examples comparing HIO ($\text{OS}=1$), HIO ($\text{OS}=2$), DAPS, FANO, and the reference. At roughly 300\,s per image, FANO reaches perceptual quality comparable to HIO at $\text{OS}=2$, recovering clean structures without any oversampling.}
\label{fig:pr_tradeoff}
\end{figure*}

Classical phase retrieval methods, such as HIO \citep{fienup1982phase}, already provide reliable recovery in the noiseless setting when the oversampling ratio satisfies $\text{OS}\ge 2$. Oversampling reduces ill-conditioning by increasing the measurement dimension, but lower oversampling ratios impose fewer requirements on optical systems and reduce acquisition costs in practice. A potential advantage of diffusion generative priors is their ability to compensate for missing oversampling information and enable high-quality recovery even at $\text{OS}=1$, without oversampling.

Figure~\ref{fig:pr_tradeoff} compares FANO, DAPS, and classical HIO on 100 FFHQ test images. The experiment follows the phase retrieval configuration in Section~\ref{subsec:exp_setup}, changing only OS to $1$ and removing measurement noise. At its largest budget of 437 seconds per image, DAPS reduces LPIPS to only approximately $0.45$, whereas FANO reaches $0.26$ at approximately 300 seconds, comparable to HIO at OS${}=2$. Representative examples in Figure~\ref{fig:pr_tradeoff}(b) further corroborate this gap: at OS${}=1$, FANO recovers structures and textures close to the reference, whereas DAPS exhibits substantial distortions and HIO fails completely at OS${}=1$.

HIO is implemented using Fourier-magnitude projections and a known image-support constraint, processing the three RGB channels separately and clipping the final output to $[0,1]$. For all $\text{OS}\in\{1,1.5,2\}$, the feedback parameter is fixed at $\beta=0.9$, and each candidate undergoes $500$ updates.

\clearpage
\section{Additional Qualitative Results}
\label{app:qualitative}
\suppressfloats[t]

We present representative reconstructions for each task below. Each figure contains two tasks (left and right); for each task, the upper half shows FFHQ and the lower half ImageNet, with three examples per dataset. The columns, from left to right, show the degraded measurement (Measurement), DAPS, SITCOM, FANO (ours), and the ground truth (Reference).

\begin{figure*}[!ht]
\centering
\setlength{\abovecaptionskip}{5pt}
\includegraphics[width=0.95\textwidth]{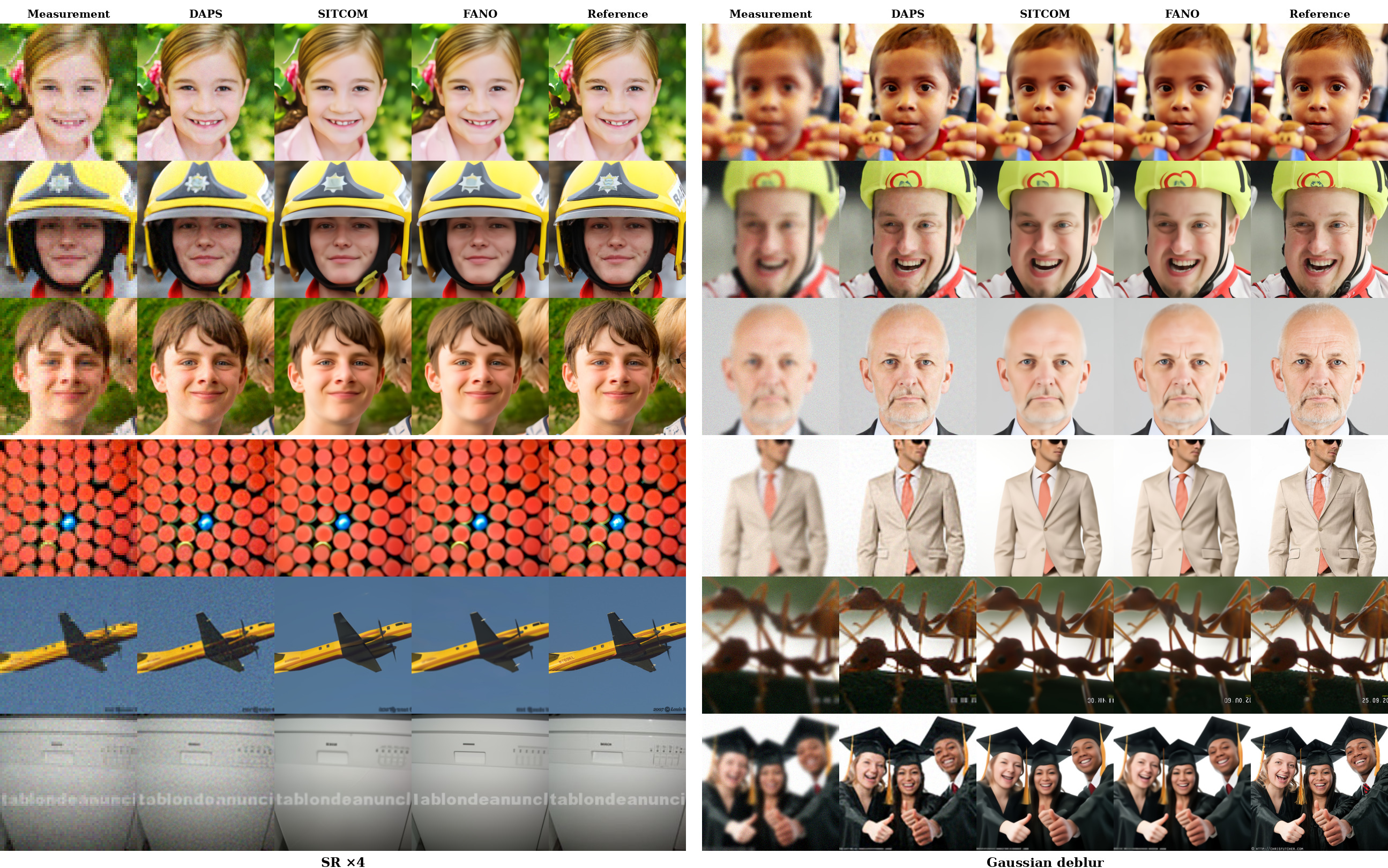}
\caption{\textbf{Super resolution} (\textit{left}) and \textbf{Gaussian deblurring} (\textit{right}) results. First (resp.\ last) three rows are for the FFHQ (resp.\ ImageNet) dataset.}
\label{fig:gallery_sr_gb}
\par\vspace{10pt}
\includegraphics[width=0.95\textwidth]{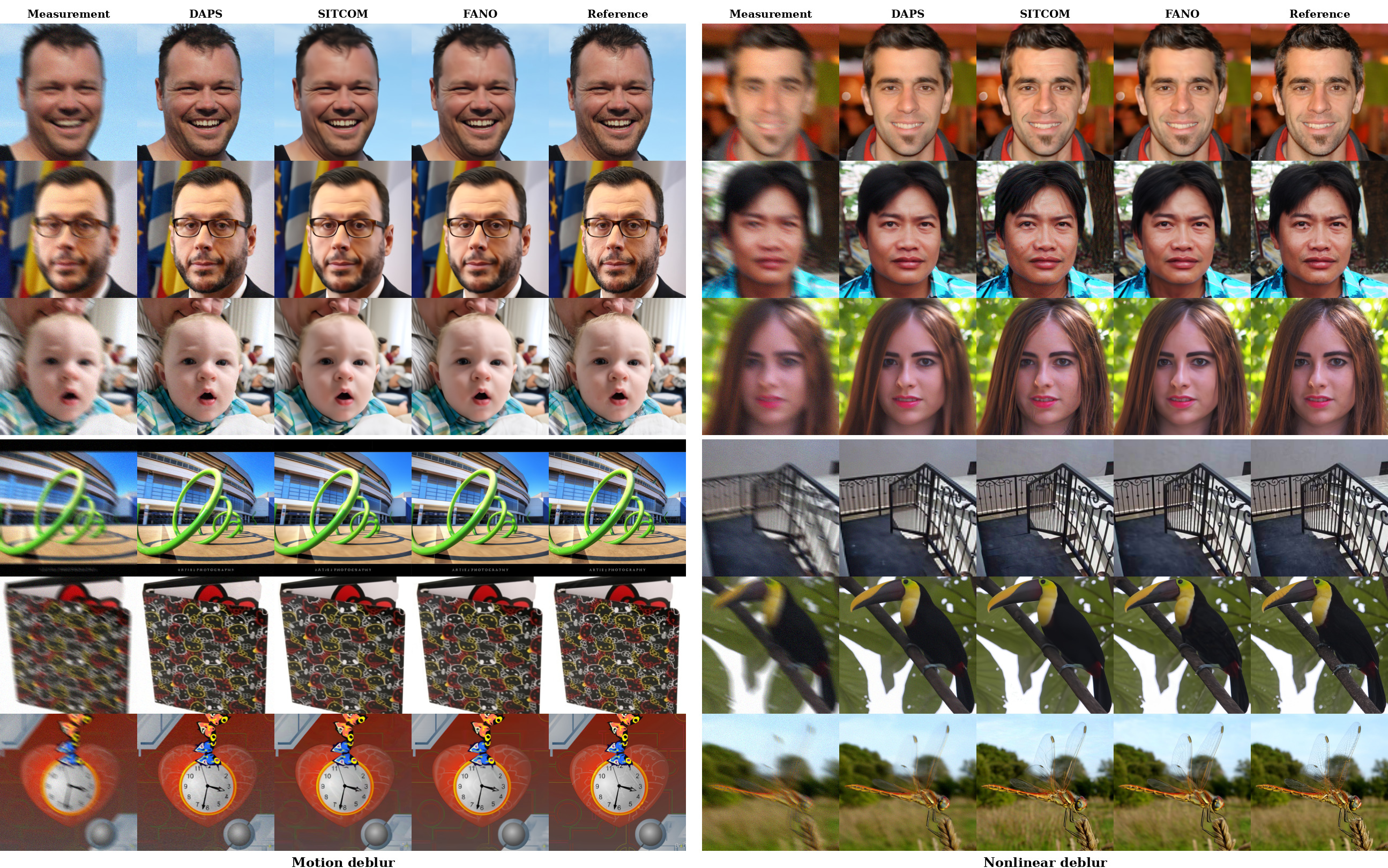}
\caption{\textbf{Motion deblurring} (\textit{left}) and \textbf{nonlinear deblurring} (\textit{right}) results. First (resp.\ last) three rows are for the FFHQ (resp.\ ImageNet) dataset.}
\label{fig:gallery_mb_nlb}
\end{figure*}

\clearpage
\begin{figure*}[!ht]
\centering
\setlength{\abovecaptionskip}{5pt}
\includegraphics[width=0.95\textwidth]{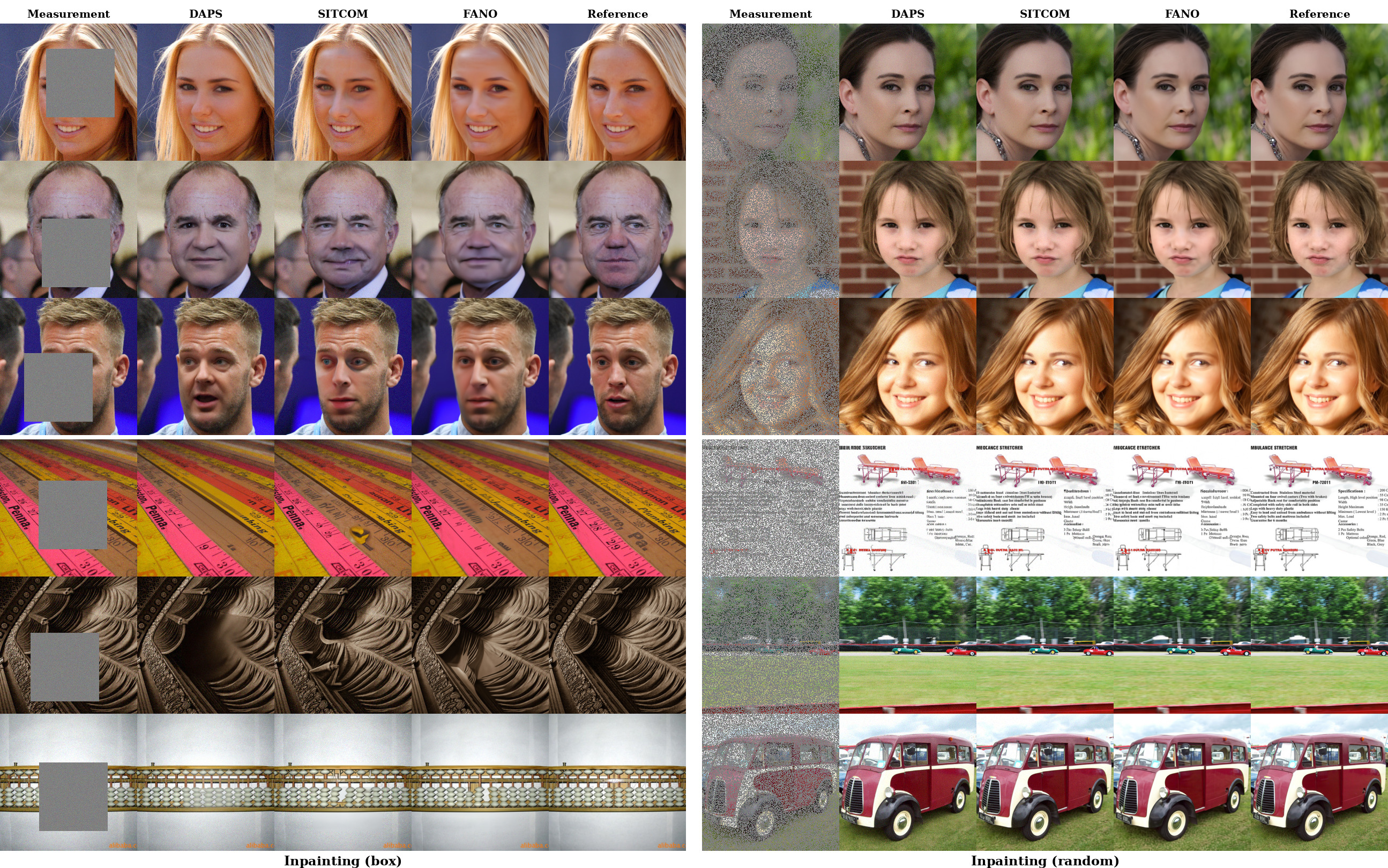}
\caption{\textbf{Box inpainting} (\textit{left}) and \textbf{random inpainting} (\textit{right}) results. First (resp.\ last) three rows are for the FFHQ (resp.\ ImageNet) dataset.}
\label{fig:gallery_inp}
\par\vspace{10pt}
\includegraphics[width=0.95\textwidth]{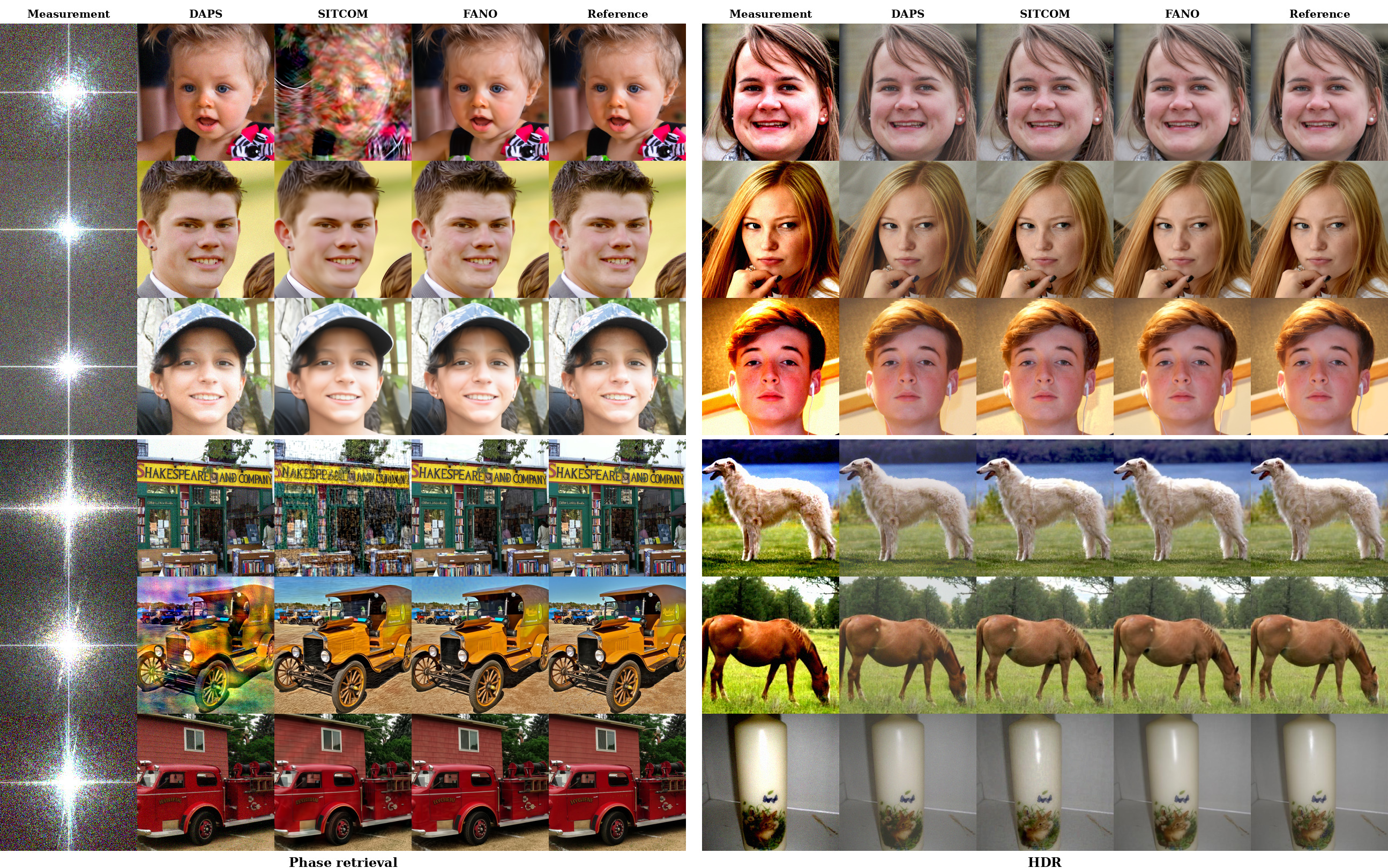}
\caption{\textbf{Phase retrieval} (\textit{left}) and \textbf{high dynamic range} (\textit{right}) results. First (resp.\ last) three rows are for the FFHQ (resp.\ ImageNet) dataset.}
\label{fig:gallery_pr_hdr}
\end{figure*}

\end{document}